\RequirePackage[svgnames]{xcolor}

\documentclass[11pt,letterpaper]{mystyle}

\usepackage[all]{hypcap}
\usepackage[svgnames]{xcolor}
\usepackage[comma,authoryear,compress]{natbib}
\usepackage{hyperref}[citecolor=lightblue]

\hypersetup{
    colorlinks = true,
    citecolor = {YaleBlue},
}

\usepackage{algorithm}
\usepackage{algorithmicx}
\usepackage{algpseudocode}
\usepackage{microtype}
\usepackage{graphicx}
\expandafter\def\csname ver@subfig.sty\endcsname{}
\usepackage{booktabs} %
\usepackage{float}
\usepackage{bigstrut}
\usepackage{amsmath}
\usepackage{wrapfig}
\usepackage{amsmath}
\usepackage{amssymb}
\usepackage{mathtools}
\usepackage{amsthm}
\usepackage{mathrsfs}
\usepackage{nicefrac}
\usepackage{dsfont}
\usepackage{enumitem}
\usepackage{subcaption}
\usepackage{graphicx,subfig}
\usepackage{cleveref}
\usepackage{bxcoloremoji}

\usepackage{float}

\usepackage[utf8]{inputenc} %
\usepackage[T1]{fontenc}    %
\usepackage{hyperref}       %
\usepackage{url}            %
\usepackage{booktabs}       %
\usepackage{amsfonts}       %
\usepackage{nicefrac}       %
\usepackage{microtype}      %
\usepackage{graphicx}
\usepackage{subcaption} 
\usepackage{amssymb}
\usepackage{fdsymbol}
\usepackage{wrapfig}
\usepackage{lipsum}
\usepackage{enumitem}
\usepackage{stackengine}
\usepackage[font=small,labelfont=bf]{caption}
\usepackage{color}
\usepackage{adjustbox}

\usepackage{rotating}
\usepackage{makecell}

\usepackage{multirow}

\newcommand{\x}{\mathbf{x}}

\definecolor{blanchedalmond}{rgb}{1.0, 0.92, 0.8}
\definecolor{carmine}{rgb}{0.59, 0.0, 0.09}
\definecolor{lightblue}{rgb}{0.22,0.45,0.70}%

\renewcommand{\mathbf}{\boldsymbol}

\makeatletter
\def\Ddots{\mathinner{\mkern1mu\raise\p@
\vbox{\kern7\p@\hbox{.}}\mkern2mu
\raise4\p@\hbox{.}\mkern2mu\raise7\p@\hbox{.}\mkern1mu}}
\makeatother

\definecolor{amaranth}{rgb}{0.9, 0.17, 0.31}
\definecolor{antiquebrass}{rgb}{0.8, 0.58, 0.46}
\definecolor{antiquefuchsia}{rgb}{0.57, 0.36, 0.51}
\definecolor{chromeyellow}{rgb}{0.31, 0.47, 0.26}

\newcommand{\github}{\raisebox{-1.5pt}{\includegraphics[height=1.05em]{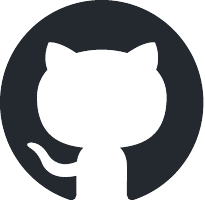}}}

\newcommand{\paperlogo}{\raisebox{-1.5pt}{\includegraphics[height=2.05em]{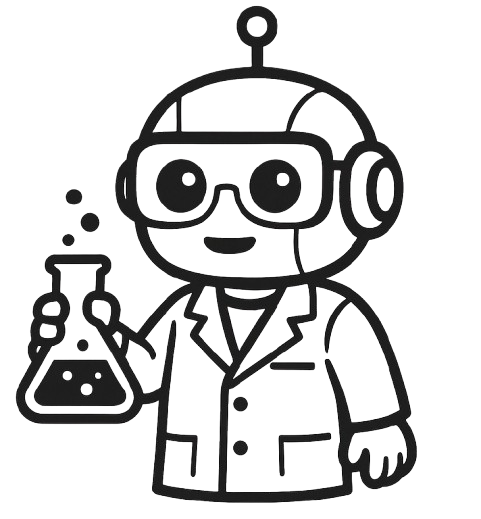}}}
\newcommand{\paperlogouniversity}{\raisebox{-1.5pt}{\includegraphics[height=2.05em]{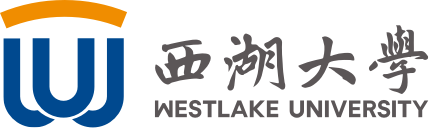}}}

\newtcolorbox{AIbox}[2][]{aibox,title=#2,#1}
\definecolor{lightblue}{rgb}{0.22,0.45,0.70}%
\definecolor{Gray}{gray}{0.95}
\definecolor{Cornsilk}{rgb}{1.0, 0.97, 0.86}

\usepackage{url}
\usepackage{hyperref}
\usepackage{xcolor}
\usepackage{pifont}
\usepackage{soul}

\usepackage{amsmath}

\usepackage[all]{hypcap}

\usepackage[utf8]{inputenc}
\usepackage[T1]{fontenc}

\usepackage{mathpazo}
\usepackage[svgnames]{xcolor} 

\usepackage{tcolorbox}
\usepackage{graphicx}
\usepackage{lipsum}

\newtcolorbox{simpleElegantQuote}{
    colback=AliceBlue!50!White,   
    colframe=RoyalBlue!75!Black,  
    boxrule=0.5pt,                
    arc=2mm,                      
    boxsep=4pt,                   
    left=10pt, right=10pt,        
    top=14pt, bottom=14pt,          
    fontupper=\itshape,           
}

\title{\paperlogo{}AI Scientists Fail Without Strong Implementation Capability}

\runningtitle{\paperlogo{} AI Scientists Fail Without Strong Implementation Capability. \quad\quad\quad\quad\quad\quad\quad\quad\quad\quad\quad\quad\quad\quad\quad\quad\quad\quad\quad\quad\quad\quad\quad\quad\quad\quad\quad\quad \paperlogouniversity{}}

\author{
  Minjun Zhu$^{1,2}$*,
  Qiujie Xie$^{1,2}$*, 
  Yixuan Weng$^1$, 
  Jian Wu$^1$, 
  Zhen Lin$^1$,
  Linyi Yang$^3$, and
  Yue Zhang
}

\affil[1]{Engineering School, Westlake University}
\affil[2]{Zhejiang University}
\affil[3]{University College London }

\correspondingauthor{Linyi Yang: \href{mailto:yanglinyiucd@gmail.com}{yanglinyiucd@gmail.com}; Yue Zhang: Email \href{mailto:zhangyue@westlake.edu.cn}{zhangyue@westlake.edu.cn}}

\begin{document}

\begin{abstract}

The emergence of Artificial Intelligence (AI) Scientist represents a paradigm shift in scientific discovery, with large language models (LLMs) taking the lead as the primary executor in the entire scientific workflow from idea generation to experiment implementation. Recent AI Scientist studies demonstrate sufficient capabilities for independent scientific discovery, with the generated research reports gaining acceptance at the ICLR 2025 workshop and ACL 2025, arguing that a human-level AI Scientist, capable of uncovering phenomena previously unknown to humans, may be imminent. Despite this substantial progress, AI Scientist has yet to produce a groundbreaking achievement in the domain of computer science on par with automated scientific tools. Based on extensive quantitative evidence from existing benchmarks in complex engineering tasks and a systematic evaluation assess 28 research papers generated by five advanced AI Scientist systems, we argue that \textbf{the fundamental bottleneck for AI Scientists lies in their capability to execute the requisite verification procedures.} Current AI Scientist systems lack the execution capabilities needed to execute rigorous experiments and produce high-quality scientific papers. To better illustrate the root cause of this \textbf{implementation gap}, we provide an in-depth discussion on the fundamental limitations of AI Scientist. This position paper aims to call for the participants in the community to bridge the implementation gap.

\vspace{5mm}

\vspace{5mm}


\vspace{5mm}

\coloremojicode{1F4C5} \textbf{Date}: May 24, 2025

\coloremojicode{1F3E0} \textbf{Projects}: \href{https://ai-researcher.net}{https://ai-researcher.net}

\github{} \textbf{Code Repository}: \href{https://github.com/ResearAI/Awesome-AI-Scientist}{https://github.com/ResearAI/Awesome-AI-Scientist}

\coloremojicode{1F4E7} \textbf{Contact}: \href{mailto:zhangyue@westlake.edu.cn}{yanglinyiucd@gmail.com, zhangyue@westlake.edu.cn}

\end{abstract}

\maketitle
\vspace{3mm}
\section{Introduction}

The automation of scientific discovery has long been one of humanity's deepest desires~\citep{langley1987scientific,king2009automation,radensky2024scideator,UniteAI2025}. In recent years, with the advances in deep neural network technology, a range of \textbf{automated scientific tools} has emerged, leading to groundbreaking achievements in fields such as biomedicine~\citep{yang2025shennongalpha,jumper2021highly}, chemistry~\citep{stokes2020deep}, and materials science~\citep{szymanski2023autonomous}. For instance, DeepMind’s AlphaFold can determine the 3D structures of proteins in just a few hours, a task that previously took years to solve~\citep{jumper2021highly}. In recent, researchers developed an autonomous laboratory, A-Lab, which successfully synthesizes 41 novel inorganic materials within 17 days~\citep{szymanski2023autonomous}.
However, these scientific tools still rely heavily on human involvement. Researchers must first formulate ideas to be tested, while AI is responsible for the labor-intensive tasks of verification and iterative search. Therefore, these systems cannot be considered as truly automated scientific research.

\begin{figure}[t]
    \centering
    \includegraphics[width=\textwidth]{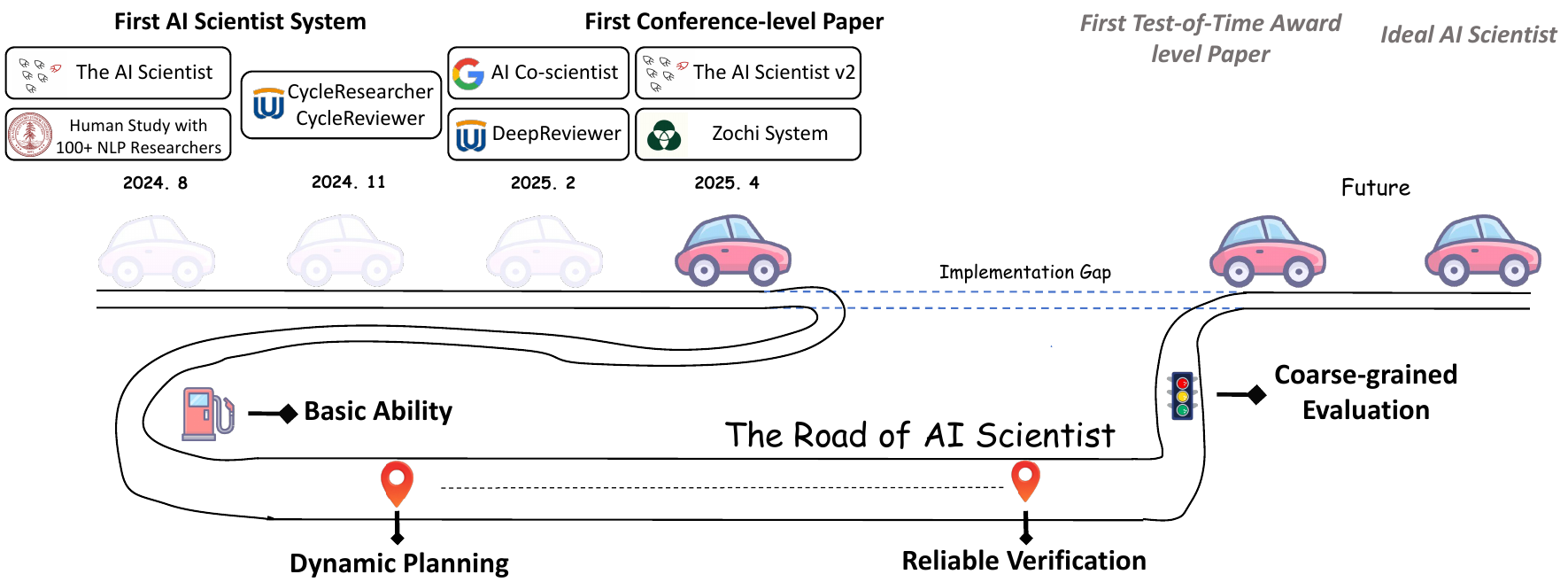}
    
    \caption{The roadmap of AI Scientist from 2024 to future, highlighting key milestones and fundamental challenges that must be overcome to bridge the implementation gap of AI Scientist.}
    \label{fig:why}
\end{figure}

The emergence of LLM-based \textbf{AI Scientist} has propelled the automation of scientific research to the next level, with AI taking the lead as the primary executor of scientific discovery, managing the entire workflow from idea generation to experiment execution~\citep{lu2024ai, weng2025cycleresearcher}. Recent studies have shown that research papers produced by AI Scientist have already reached the level of submissions to major machine learning conferences~\citep{si2024can,yamada2025ai,zochi2025}. As shown in Figure~\ref{fig:why}, we demonstrate the progress made by AI Scientist-v2~\citep{yamada2025ai}, and the research output has received review scores exceeding the average acceptance threshold for human-authored papers. Similarly, researchers present an empirical validation through multiple peer-reviewed publications accepted at ICLR 2025 workshops and ACL 2025 main conference~\citep{zochi2025}. Despite this substantial progress, AI Scientist has yet to produce a groundbreaking achievement in the domain of computer science on par with automated scientific tools~(e.g., AlphaFold~\citep{jumper2021highly}).

In this position paper, we first propose a conceptual framework (Section \ref{sec:definition}) that defines an AI Scientist as \textbf{an advanced end-to-end system capable of independently formulating scientific ideas and performing the implementation for verifying these ideas.} This definition forms the theoretical foundation of our position, aligns with current research progress~\citep{lu2024ai,weng2025cycleresearcher,yamada2025ai}, and emphasizes that the core capability of an AI Scientist lies in generating innovative and feasible ideas at scale~\citep{si2024can, wang-etal-2024-scimon, hu2024nova, yang2025moosechem}. The idea-generation capability is a key feature that sets AI Scientists apart from automated scientific tools. While recent advances demonstrate that AI Scientists can generate highly innovative ideas \citep{SiYH25}, their implementation capabilities remain constrained \citep{chan2024mle,starace2025paperbench,xiang2025scireplicate,siegel2024core,mldevbench}, creating a significant gap between innovative idea generation and complete implementation.

\begin{simpleElegantQuote}
\textbf{Our Position:} The fundamental bottleneck for AI Scientists lies in their implementation capability to effectively execute the verification of these ideas.
\end{simpleElegantQuote}

We defend our argument by analyzing quantitative evidence from existing benchmarks used to evaluate LLMs' abilities in performing complex engineering tasks~(Section \ref{subsec:benchmark_results}). While LLMs can generate highly novel ideas~\citep{si2024can,chai2024exploring,gottweis2025towards}, their performance in experiment execution is exceptionally poor~(Table~\ref{tab:mod_benchmarks}). For instance, a leading LLM like Claude 3.5 Sonnet scored only \textbf{1.8\%} on PaperBench~\citep{starace2025paperbench}. This \textbf{implementation gap} is further supported by a systematic evaluation~(Section \ref{subsec:peer_review_of_ai_papers}), which leverages a state-of-the-art review model, DeepReviewer-14B~\citep{zhu2025deepreview}, to assess 28 research papers generated by five advanced AI Scientist systems. The results demonstrate that current AI Scientist systems lack the execution capabilities needed to execute rigorous experiments and produce high-quality scientific papers. Finally, to clearly illustrate the root cause of the implementation gap, we provide an in-depth discussion on the fundamental limitations of AI Scientist~(Section \ref{sec:limitations}).

In summary, this paper validates and deeply analyzes the implementation gap in existing AI Scientist systems based on extensive quantitative evidence and a simulated peer-review process. Furthermore, as the development of AI Scientists will bring greater regulatory challenges, we comprehensively examine the ethical considerations~(Section \ref{sec:ethical_consideration}) faced by AI Scientists and suggest directions for future research~(Section \ref{sec:future}). We hope this position paper will contribute to a clearer understanding of the limitations of current AI Scientist, shedding light on the future development of AI Scientist.

\section{Definition of the AI Scientist}
\label{sec:definition}


The emergence of automated scientific tools has accelerated scientific discovery across numerous domains~\citep{king2009automation, yang2025shennongalpha,jumper2021highly,stokes2020deep,szymanski2023autonomous}. However, these tools fundamentally operate within a paradigm where human researchers remain in the dominant position of scientific discovery, and thus cannot be classified as fully automated AI Scientists. In this section, we first provide a detailed discussion of the unique characteristics of the AI Scientist~(Section \ref{subsec:characteristics}). Building on this discussion, we then propose a conceptual framework that formally defines the AI Scientist in a mathematical form~(Section \ref{subsec:framework}).

\subsection{Unique Characteristics}
\label{subsec:characteristics}

\newlength{\oldintextsep}
\setlength{\oldintextsep}{\intextsep}
\setlength{\intextsep}{0pt}
\setlength{\columnseprule}{10pt}
\begin{wrapfigure}{t}{0.58\textwidth}
    \centering
    \includegraphics[width=1\linewidth]{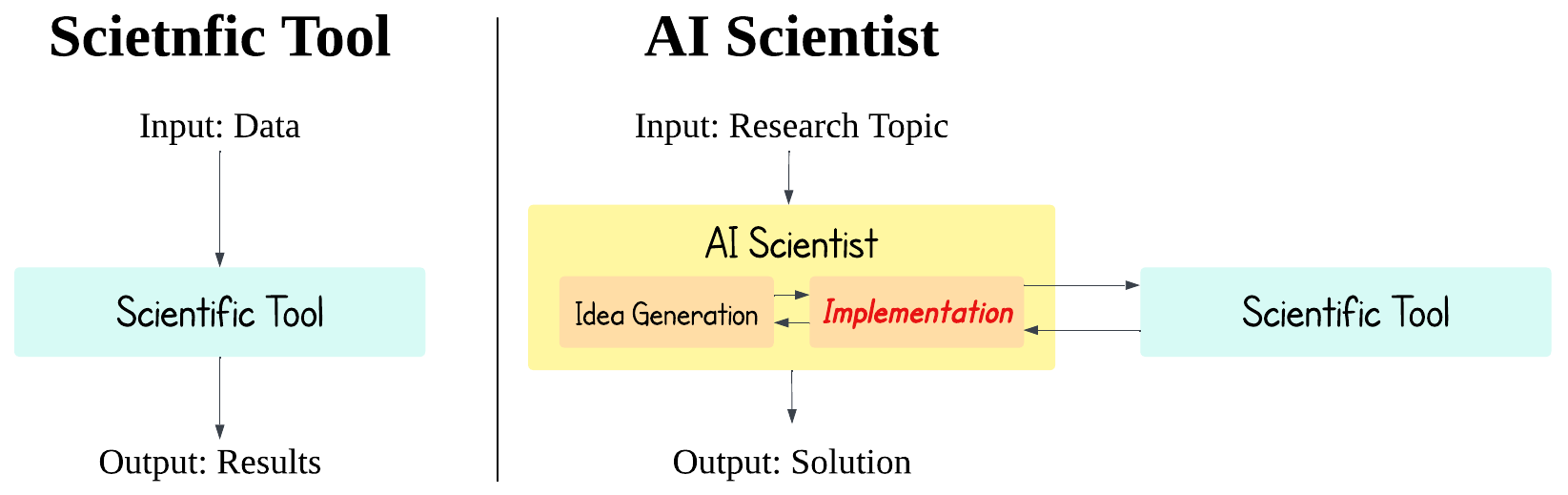}
    \vspace{-0.3cm}
    \caption{The difference between AI Scientist and scientific tool. 
    Scientific tools generate predictions under human supervision, while AI Scientist autonomously leverage tools to address research questions.
    }
    \label{fig:difference}
\end{wrapfigure}

Scientific tools, originating from AI for Science research, represent specialized AI systems designed to solve specific scientific problems by processing data and generating results within defined domains. These tools have demonstrated remarkable success across diverse scientific fields, including protein structure prediction (e.g. AlphaFold) \citep{jumper2021highly}, antibiotic discovery through deep learning approaches~\citep{stokes2020deep}, and autonomous chemical research with large language models~\citep{boiko2023autonomous}. These scientific tools fundamentally operate within a knowledge-dependent collaborative framework between humans and AI.

AI Scientist represents a research paradigm shift where AI assumes the role of an autonomous scientist capable of conducting independent scientific research. As illustrated in Figure~\ref{fig:difference}, while scientific tools operate under human supervision, receiving data as input and producing predictions as output, AI Scientist goes a step further by demonstrating autonomous scientific reasoning capabilities. It accepts research questions as input and engages in iterative, self-directed interactions with scientific tools to generate comprehensive solutions. Unlike scientific tools that function as sophisticated instruments awaiting human guidance, AI Scientist exhibits genuine scientific agency, conducting end-to-end scientific investigations from question formulation to solution discovery~\citep{yamada2025ai}.

\subsection{Conceptualized Framework}
\label{subsec:framework}
\begin{simpleElegantQuote}
\textbf{Our Definition:} An AI Scientist is an advanced end-to-end system capable of independently formulating scientific ideas and executing the requisite verification and falsification procedures.
\end{simpleElegantQuote}

We define an AI Scientist, denoted as~$\mathcal{S}_{AI}$, as a fully autonomous scientific intelligence capable of independently performing diverse scientific research tasks. Different from a general scientific tool, it must possess dual capacities, including idea generation and experimental execution.
A complete scientific research task typically originates from an initial scientific question $\mathcal{Q}_{init}$ and leverages existing domain knowledge $\mathcal{K}_{domain}$. An AI Scientist, denoted $\mathcal{S}_{AI}$, operates within the scope of human ethical constraints $\mathcal{R}_{human}$ and resource constraints $\mathcal{B}_{res}$ to conduct this task. The primary output is the generation of novel scientific knowledge $\mathcal{K}_{new}$ and associated verifiable artifacts $\mathcal{A}_{sci}$. The process through which an AI Scientist aims to achieve the optimal output from a scientific research task can be formally represented as:

\begin{equation}(\mathcal{K}_{new}, \mathcal{A}_{sci}) \leftarrow \max  \{\mathcal{S}_{AI}(\mathcal{Q}_{init},\mathcal{K}_{domain}, \mathcal{R}_{human} |\theta_{AI}, \mathcal{B}_{res})\}\end{equation}

\setlength{\intextsep}{\oldintextsep}

\section{Arguments for Implementation Capability}
\label{sec:arguments}



\textbf{We argue that the fundamental bottleneck limiting AI Scientists lies not in their idea generation capabilities, but in their capacity to execute rigorous implementation procedures required for reliable scientific research.} To support this position, we present three lines of evidence: systematic analysis of research trends in the AI Scientist literature (Section \ref{subsec:research_trends}), comprehensive benchmark analysis across multiple evaluation frameworks (Section \ref{subsec:benchmark_results}), and systematic peer review assessment using LLM-as-a-Judge methodology (Section \ref{subsec:peer_review_of_ai_papers}).

\subsection{Research Trend of AI Scientist}
\label{subsec:research_trends}

\begin{wrapfigure}{t}{0.48\textwidth}
\resizebox{0.47\textwidth}{!}{
\begin{tabular}{ccc}
\toprule
\textbf{Type} & \textbf{Total citations}  & \textbf{Avg. citations} \\
\midrule
Without Implementation & 216 & 10.3 \\
\rowcolor[rgb]{ .949,  .949,  .949}
With Implementation & \textbf{325}  & \textbf{25.0} \\
\bottomrule
\end{tabular}}

    \centering
    \includegraphics[width=0.95\linewidth]{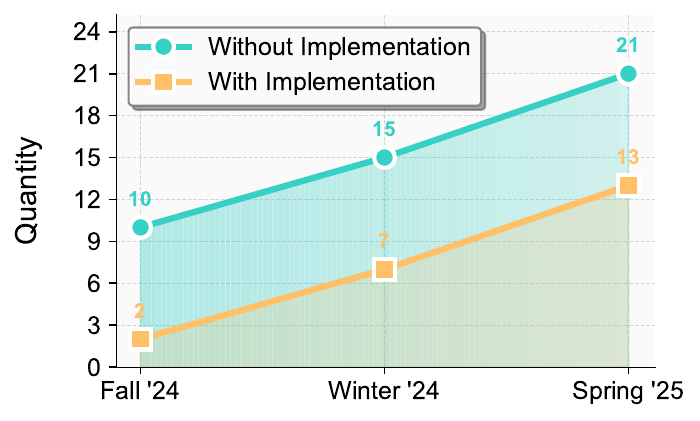}
    \caption{Analysis of AI Scientist publications on arXiv. The upper panel displays the average number of citations up to now, categorized by containing implementation details. The lower panel shows the growth in the total number of these papers with the same categorization.}
    \label{fig:difference_im}
\end{wrapfigure}

Our statistical analysis of AI Scientist papers on arXiv up to May 23, 2025 (see Appendix~\ref{sec:statistics} for details), reveals key trends illustrated in Figure~\ref{fig:difference_im}. The lower panel of the figure shows that while the total number of publications is growing, studies focusing on idea generation without concrete implementation details consistently outnumber those incorporating such implementations. Despite this disparity in publication numbers, the upper panel indicates a crucial counterpoint: papers that include substantive implementation details achieve a significantly higher average number of citations. This signals strong community valuation for executable advancements and underscores the importance of addressing the implementation gap. This then raises a critical question: if implementation-focused research garners higher impact, why does its volume remain markedly lower? This disparity strongly implies that the path of implementation is fraught with substantial challenges.

\setlength{\intextsep}{\oldintextsep}
\begin{table}[t]
\centering
\small 
\caption{State-of-the-art (SoTA) LLMs show relatively low accuracy on code implementation on different tasks. The listed benchmarks are collected from diverse domains. The table below details their tasks, domains, scale, methods, and performance.} 
\label{tab:mod_benchmarks}
\resizebox{\textwidth}{!}{ 
\begin{tabular}{@{} r p{4.2cm} p{3.2cm} r r r @{}} 
\toprule
\textbf{Benchmark} & \textbf{Task Description} & \textbf{Domains} & \textbf{Scale} & \textbf{LLM} & \textbf{Acc. Performance} \\
\midrule

MLE-Bench\citep{chan2024mle} &
AI Training task & 
Applied ML&
75&
OpenAI o1-preview &
16.90\% \\

\rowcolor[rgb]{ .949,  .949,  .949}
PaperBench \citep{starace2025paperbench}&
ICML paper Replicating& 
NLP, CV, ML&
8,316&
OpenAI o1-high &
26.00\%\\

SciReplicate-Bench \citep{xiang2025scireplicate}&
Code Generation & 
NLP &
100&
Claude-Sonnet-3.7 &39.00\%\\

\rowcolor[rgb]{ .949,  .949,  .949}
CORE-Bench \citep{siegel2024core}&
Scientific Paper reproduction & 
Computer Science, Social Science, and Medicine&
270 &
OpenAI GPT-4o  &
55.56\% \\

ML-Dev-Bench \citep{mldevbench}&
AI training task & 
ML &
30 &
Claude-Sonnet-3.5&
50.00\% \\
\bottomrule

\end{tabular}}
\end{table}

\subsection{Quantitative Analysis}
\label{subsec:benchmark_results}

\textbf{Empirical Evidence of Implementation Gap.} 
Advanced LLMs achieve near-saturated performance on simple code generation benchmarks like HumanEval \citep{chen2021evaluating,liu2023your,yang2025qwen3}. For example, o3 exhibits excellent problem-solving capabilities in the 99.8th percentile of human performance on algorithmic competition platforms like Codeforces.
However, the performance of SoTA LLMs \textbf{drops dramatically when it comes to real-world research scenarios}. As depicted in Table~\ref{tab:mod_benchmarks}, we compare existing benchmarks for evaluating LLM agents' abilities to perform challenging machine learning research tasks, including MLE-Bench~\citep{chan2024mle} (solving Kaggle machine learning tasks, evaluated by medal rates), PaperBench~\citep{starace2025paperbench} (replicating ICML research papers from scratch, assessed by replication scores), SciReplicate-Bench~\citep{xiang2025scireplicate} (generating executable code from NLP algorithm descriptions, measured by execution accuracy), CORE-Bench~\citep{siegel2024core} (reproducing computational results from scientific papers, determined by accuracy), and ML-Dev-Bench~\citep{mldevbench} (completing diverse ML development workflow tasks, assessed by success rates). Each takes a different approach to measuring how well AI systems can automate aspects of ML research. These evaluations consistently demonstrate that LLMs face difficulty in translating conceptual understanding or initial plans into verifiably correct and operational code. \textbf{This ``implementation gap'' fundamentally limits AI Scientist's verification capabilities.}


\textbf{Beyond Code Generation.} The complexity of real-world research implementation processes extends far beyond simple code generation tasks, often requiring sustained reasoning and multi-step problem-solving. However, current LLMs exhibit relatively weak performance on such complex challenges. LiveCodeBench (LCB) \citep{jain2024livecodebench}, a more complex evaluation benchmark than Humaneval \citep{chen2021evaluating} that collects problems from periodic contests on LeetCode, AtCoder, and Codeforces platforms, evaluates Code LLMs across diverse code-related scenarios, including code generation, execution, self-repair, and output prediction. o4-mini achieves SoTA performance on the code generation subtask with \textbf{only 52.1$\%$ pass@1 score}. \textbf{This poor performance on complex coding tasks reveals that AI scientists lack the implementation ability to handle sophisticated code-based research scenarios.} 


\textbf{Implementation and verification.} We observe that the verification bottleneck emerges across multiple stages of the research process. SciReplicate-Bench~\citep{xiang2025scireplicate}, which tasks LLM agents with generating Python code to reproduce algorithms from NLP research papers, reveals that despite agents demonstrating an understanding of algorithmic logic (evidenced by high reasoning graph accuracy), they struggle with code execution. The best agent achieved only \textbf{39$\%$} execution accuracy, indicating its generated code passed functional test cases for just \textbf{39$\%$} of the tasks, highlighting \textbf{a failure to ensure implementation correctness and runtime behavior.} Similarly, PaperBench~\citep{starace2025paperbench} requires LLM agents to replicate entire machine-learning papers from scratch by developing codebases and running experiments. While agents can generate code components (e.g., o1-High achieving \textbf{43.4$\%$} success on weighted "Code-Development" sub-tasks), their performance on subsequent verification stages is poor. On rubric-defined leaf nodes for ``Execution'' (successfully running the code) and ``Result Match'' (quantitatively matching the paper's reported results), Claude 3.5 Sonnet scored only \textbf{1.8$\%$} and \textbf{0.7$\%$} respectively. This poor performance indicates a breakdown in ensuring the developed solution operates correctly and produces the intended outcomes.

\textbf{Discussion.} The verification challenge extends beyond initial code implementation to debugging, iterative refinement, and validation of experimental outcomes. Evidence from MLE-Bench and ML-Dev-Bench~\citep{chan2024mle, mldevbench} shows that LLM agents frequently fail to debug their code or produce valid submissions, with \textbf{20\%} of o1 preview runs on MLE Bench failing this step, and struggle to optimize model performance. Debugging, an explicit verification procedure, also indicates persistent agent failures that highlight the verification bottleneck \citep{chan2024mle}. The incapacity to iteratively refine solutions towards better performance, illustrated in ML-Dev-Bench where all tested agents scored \textbf{0\%} on ``Model Performance'' tasks, further signifies deficiencies in robust verification loops essential for scientific advancement \citep{mldevbench}. Furthermore, CORE-Bench, which requires agents to reproduce results and then answer questions based on these outputs, assesses the verification of entire computational experiments. This process, involving multiple stages of reproduction and reasoning, presents significant challenges. For instance, the imperfect success rates (e.g., CORE-Agent with GPT-4o achieved \textbf{55.56\%} on CORE-Bench Medium) highlight the difficulties in this complex verification process \citep{siegel2024core}. These difficulties across verification tasks suggest that enhancing AI Scientists' systematic verification capability is crucial for their maturation into ideal AI Scientists. Current LLMs, while proficient in content generation, fail to rigorously validate their outputs against explicit criteria, a foundational component of scientific practice.


\subsection{LLM-as-a-Judge Reveals the Implementation Weaknesses}
\label{subsec:peer_review_of_ai_papers}
To further support the existence of implementation gap, we employ a simulated peer review methodology to assess the actual quality of scientific outputs from current AI Scientist systems, particularly their implementation-level reliability.
We select 28 publicly available research papers generated independently by five different AI Scientist systems and utilize the SoTA review model DeepReviewer-14B \citep{zhu2025deepreview} to conduct systematic evaluation under unified standards. We acknowledge that potential selection bias in the public availability of these papers (e.g., researchers may only publish better-performing outputs) means our evaluation results may not fully represent the average output quality of these systems across all scenarios. Nevertheless, \textbf{this analysis provides valuable insights into the general quality level of current AI-generated research papers.}

\begin{table}[t]
\centering
\scriptsize
\caption{DeepReviewer-14B Evaluation of AI-Generated Papers from Various AI Scientist Systems. Scores reflect averages across the 'Num' of available papers. Note: Publicly available papers may be curated and not fully representative of typical system output.}

\label{tab:ai_scientist_performance_why}

\begin{tabular}{cccccccc}
\toprule
\textbf{AI Scientist System} & \textbf{Num}  & \textbf{Soundness$\uparrow$} & \textbf{Presentation$\uparrow$} & \textbf{Contribution$\uparrow$} & \textbf{Decision$\uparrow$} & \textbf{Rating$\uparrow$}& \textbf{Percentile$\uparrow$}\\
\midrule
HKUSD AI Researcher & 7 & 1.75 & 1.46 & 1.57 & 0.0 & 2.57 &3.43\%\\
\rowcolor[rgb]{ .949,  .949,  .949}
AI Scientist & 10  & 2.08 & 1.80 & 1.75& 0.0 & 3.35&8.22\% \\
AI Scientist v2 & 3 & 1.67 & 1.50 & 1.50& 0.0 & 2.33 &2.04\% \\
\rowcolor[rgb]{ .949,  .949,  .949}
CycleResearcher-12B & 6  & 2.25 & 1.75 & 2.13& 0.0 & 3.75& 16.88\%\\
Zochi & 2  & 2.38 & 2.38 & 2.25& 0.0 & 4.63&29.96\% \\
\bottomrule
\end{tabular}
\end{table}

\setlength{\intextsep}{5pt}
\setlength{\columnseprule}{10pt}
\begin{wraptable}{r}{0.46\textwidth}
  \centering
  \caption{Defect Categories and Their Issues.}
  \label{tab:defect_categories}
  \small
\resizebox{0.45\textwidth}{!}{
  \begin{tabular}{l r r}
    \toprule
    \textbf{Defect Category} & \textbf{Number} & \textbf{Percentage} \\
    \midrule
    Experimental Weakness & 28 & 100\% \\
    \rowcolor[rgb]{ .949,  .949,  .949}
    Methodological Unclarity/Flaws & 27 & 96.4\% \\
    Writing \& Presentation Issues & 26 & 92.9\% \\
    \rowcolor[rgb]{ .949,  .949,  .949}
    Novelty Concerns & 25 & 89.3\% \\
    Theoretical Weakness & 24 & 85.7\% \\
    \rowcolor[rgb]{ .949,  .949,  .949}
    Literature Review Deficiencies & 22 & 78.6\% \\
    Practicality \& Robustness Gaps & 21 & 75.0\% \\
    \rowcolor[rgb]{ .949,  .949,  .949}
    Reproducibility Issues & 20 & 71.4\% \\
    Computational Cost Concerns & 18 & 64.3\% \\
    \rowcolor[rgb]{ .949,  .949,  .949}
    Component Analysis & 16 & 57.1\% \\
    Hyperparameter Analysis Lacking & 16 & 57.1\% \\
    \rowcolor[rgb]{ .949,  .949,  .949}
    Ethical Considerations Missing & 3 & 10.7\% \\
    \bottomrule
  \end{tabular}}
\end{wraptable}

\textbf{Quantitative Assessment Results.} As Table~\ref{tab:ai_scientist_performance_why} demonstrates, DeepReviewer-14B assigns generally low average scores to these AI-generated papers across multiple core dimensions. The Rating scale ranges from 1-10, where a score of 6 indicates acceptable quality, while Soundness, Presentation, and Contribution scores range from 1-4. The highest-rated system, Zochi (with a sample size of 2 papers), achieves \textbf{an average rating of only 4.63}, while other systems score even lower, typically in the 2-3 point range. These quantitative scores reveal that current AI Scientist systems face \textbf{significant challenges in independently producing high-quality research papers.}

Table~\ref{tab:defect_categories} shows that among the twelve major defect categories, ``Experimental Weakness'' appears across all 28 evaluated AI-generated papers, with a 100$\%$ occurrence rate. \textbf{This finding supports our positions regarding implementation capability limitations}, in experimental design, execution, and result analysis. The second and third most prevalent issues are ``Methodological Unclarity/Flaws'' (96.4$\%$) and ``Writing \& Presentation Issues'' (92.9$\%$), which reflect AI Scientists' insufficient ability to clearly articulate and implement research plans. ``Novelty Concerns'' (89.3$\%$) and ``Theoretical Weakness'' (85.7$\%$) occur frequently, indicating that when AI Scientists generate complete papers, they struggle to propose original scientific contributions with solid theoretical foundations. The prevalence of these high-frequency defects highlights \textbf{systemic issues in the scientific rigor and implementation quality of current AI-generated research}, falling below the standards for reliable and valuable scientific outputs.

\setlength{\intextsep}{\oldintextsep}
\section{Rooted Limitations of Execution Capabilities}
\label{sec:limitations}

Our empirical analysis~(Section \ref{sec:arguments}) reveals a clear pattern that while AI Scientists are conceptualized as advanced iterations of traditional scientific tools, they consistently fail at implementation and verification procedures across diverse scientific contexts. 
This raises a critical question:  Why do these sophisticated systems fail to achieve consistently strong results, especially when traditional scientific tools, wielded by human researchers, prove highly effective? 
To understand this paradox, we provide a discussion on the root cause of the implementation gap~(Section \ref{subsec:root_cause} ) and present an in-depth analysis of the fundamental limitations of AI Scientist~(Section \ref{subsec:rooted_limitations}).

\subsection{Two Primary Facets of Implementation Gap}
\label{subsec:root_cause}

The implementation gap for AI Scientists comprises two primary facets: (1) \textbf{AI Scientists often exhibit bottlenecks in the planning and execution stages}. This manifests in three key areas: \textbf{failures in long-range logical reasoning} required for coherent experimental design, \textbf{inadequate multi-agent collaboration} capabilities including strategic planning across complex multi-file implementations and converting conceptual ideas into working code, and \textbf{insufficient coordination} with external tools and systems; (2) Even when implementation code is generated, \textbf{AI Scientists demonstrate fundamental weaknesses in evaluation processes.} This includes failures in debugging capabilities, experimental validation, result interpretation, and iterative refinement based on experimental feedback. Current systems lack robust mechanisms for assessing implementation quality, validating experimental outcomes, and providing reliable feedback loops that can guide subsequent implementation improvements.

\textbf{Prevent building ``castle in the air''.} Agent tools often produce difficult-to-verify code and experiments, while evaluation gaps prevent AI Scientists from recognizing and correcting implementation issues through iterative refinement. Without fundamentally enhancing both capabilities, the idealized AI Scientist capable of independent scientific exploration will remain inefficient.

\subsection{Rooted Limitations}
\label{subsec:rooted_limitations}

From the current literature on AI scientists, we conclude four major limitations that collectively explain why AI scientists struggle with complex, multi-stage implementation processes: 


\textbf{Limitation 1: fundamental cognitive and execution capabilities.} Scientific implementation requires sophisticated long-range logical reasoning across multiple abstraction levels. Existing LLMs demonstrate significantly decreased coherence and robustness as reasoning chains extend \citep{wu2025more,wu2025shifting}, and increased thinking time does not necessarily yield stronger performance \citep{ballon2025relationship}. Furthermore, LLM-based agents possess limited capacity to retain past interaction information, with memory deteriorating as text length increases \citep{pink2025position,cemri2025multi}. Most critically, mainstream language models exhibit markedly weaker performance in multi-turn dialogues or multi-step interactive tasks requiring context coherence, deep understanding, and state tracking, with average performance decreases reaching 39$\%$ \citep{laban2025llms}. This capability degradation in scenarios involving long-range dependencies and complex interactions directly \textbf{constrains AI Scientist performance in executing complex scientific experiments} requiring sustained attention and coherent reasoning chains.

\begin{wrapfigure}{t}{0.68\textwidth}
\centering
\includegraphics[width=0.66\textwidth]{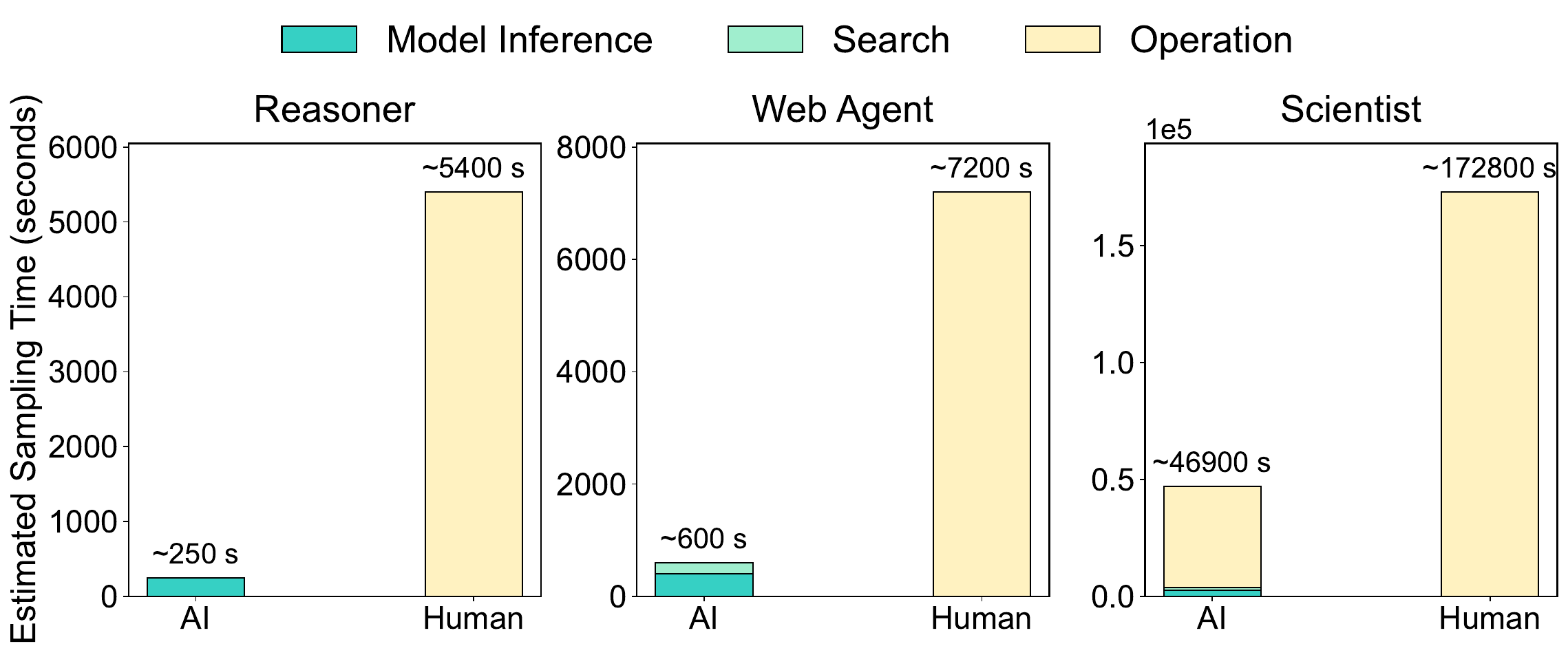}
\caption{Estimated time to solve representative tasks for different agent types (AI vs. Human), which for AI agents also corresponds to single-sample RL sampling duration. }
\label{fig:sampling_time_estimation}
\end{wrapfigure}

\textbf{Limitation 2: strategic planning and reasoning.} Scientific implementation requires comprehensive abilities for strategic reasoning, continuous monitoring, and dynamic adjustment across all research stages \citep{lu2024ai,yamada2025ai}. 
High-quality research implementation demands global planning abilities spanning entire codebases, which typically contain multiple interdependent files with hundreds of lines requiring coordinated modification \citep{SWEBenchCanLanguageModels,aleithan2024swe}. Long-term, complex scientific exploration tasks such as discovering new materials, and modeling complex biological systems particularly require continuous iteration of research directions and experimental strategies over extended time scales based on emerging results and external feedback \citep{merchant2023scaling,brixi2025genome,weng2023large}. However, current LLMs demonstrate inadequate adaptive planning and metacognitive abilities when handling highly open, creative scientific research requiring dynamic adjustments to overall research blueprints. While reinforcement learning approaches may potentially enhance LLMs' generalization and metacognitive capabilities, the resource investment required for ``inventor'' roles like AI Scientists that need to perform complex asynchronous operations and real-world interactions proves enormous. Figure~\ref{fig:sampling_time_estimation} highlights AI's acceleration over human performance in complex tasks such as reasoning and web-based research. While AI Scientists also achieve tasks faster than humans, their estimated single-sample RL training time is orders of magnitude greater than simpler AI agents. This substantial increase in required sampling time (detailed in Appendix~\ref{appendix:1}) underscores the immense challenge of developing AI Scientists via standard RL methodologies.

\textbf{Limitation 3: multi-agent collaboration.} Ideal AI Scientist should seamlessly integrate into complex research ecosystems, engaging in efficient, accurate interactions and coordination with human scientists, other AI agents, and external tools \citep{guo2024large,qian2024scaling,pu2025piflowprincipleawarescientificdiscovery}. This requires AI Scientist to not only understand instructions conforming to collaborative protocols but also precisely execute the implementation phases assigned to it within tasks and reliably feed its outputs back to the collaborative network \citep{bo2024reflective,zhang2024chain}. However, current LLM Agents still have considerable room for improvement in robustness and adaptability when interacting with dynamic environments \citep{wei2025browsecomp}. For instance, when calling a series of external APIs to complete a complex scientific computational process, LLM often struggles to handle subtle changes in API interfaces, 
and other practical engineering issues \citep{shen2025shortcutsbenchlargescalerealworldbenchmark}.

\textbf{Limitation 4: evaluation and verification.} Existing benchmarks 
such as MLE-Bench \citep{chan2024mle} and PaperBench  \citep{starace2025paperbench} primarily focus on the complete reproduction of code and experiments from papers. SciReplicate-Bench \citep{xiang2025scireplicate} emphasizes generating necessary code from scientific papers, while ScienceAgentBench \citep{chen2025scienceagentbench} concentrates on independent and singular data-driven tasks. \textbf{However, there is currently a lack of a comprehensive benchmark that can evaluate the entire scientific workflow}, from initial idea generation through to final implementation and completion. This absence makes it difficult to fairly compare the end-to-end capabilities of different AI Scientist systems. 

Additionally, there is a deficiency in evaluation approaches that incorporate measures for external supervision during the AI Scientist's implementation process. The deeper issue is that the quality of scientific discovery often lacks unified objective standards, and the process of scientific exploration is filled with uncertainty and openness, making comprehensive evaluation and effective supervision of AI Scientist's verification capability exceptionally difficult. Evaluating AI Scientist's output (e.g., generated papers) from a peer review perspective, while being a results-oriented assessment method, also has inherent limitations. As in human research systems, even experienced peer reviewers may not always accurately identify the groundbreaking and far-reaching work. A frequently cited example is that the word2vec paper \citep{mikolov2013distributed} was initially rejected by ICLR 2013, but later received the ``Test of Time Award'' at NeurIPS 2023. Extensive experimental analyses have demonstrated that review scores are not reliable indicators for predicting future impact \citep{abramo2019peer,cortes2021inconsistency}, suggesting that peer review may be more suitable for filtering low-quality papers rather than identifying the highest quality papers. 

\section{Ethical Considerations}
\label{sec:ethical_consideration}
\begin{simpleElegantQuote}

\textbf{Sub-Position:} AI scientists are in urgent need of a comprehensive system for generation management and quality evaluation.

\end{simpleElegantQuote}

As autonomous research agents, AI Scientists lack values and moral constraints. They are incapable of making ethical judgments about the societal impact of their work, and they do not self-regulate based on potential risks associated with their findings~\citep{bengio2025superintelligent}. As AI Scientists possess stronger capabilities in idea generation and experiment execution, their influence on scientific research and society could far surpass that of current LLMs and scientific tools (e.g., Deep Search, AutoSurvey~\citep{wang2024autosurvey}). In the absence of proper oversight, \textbf{AI Scientists may: (1) be misused}, overwhelming the peer review system, leading to a decline in overall research quality; \textbf{(2) enter unethical or dangerous research domains}, autonomously generating and publishing sensitive findings that accelerate the development of harmful technologies; \textbf{(3) weaken the quality of PhD training}, leading to a decline in human research standards and overall scientific literacy. To prevent the above situations, \textbf{we argue that AI Scientists are in urgent need of a comprehensive system for generation management and quality evaluation}, thus enabling effective behavior regulation within the human moral framework~\citep{jobin2019global}. This system should include, but not be limited to, the following components:

\textbf{(1) Implement measures to prevent AI-generated content from disrupting human review systems:} Effective strategies should be adopted to ensure that AI-generated articles do not interfere with human peer-review systems while maintaining high standards of quality. This includes establishing a centralized platform to archive scientific outputs generated by AI Scientists, developing automated detection systems to identify such content, and creating specialized evaluation tools (e.g., DeepReview~\citep{zhu2025deepreview}) to assess the quality of AI-generated research outputs. These tools should help identify and filter low-quality content, thereby reducing the burden on the peer review process. All AI-generated outputs must be transparently labeled and reviewed, including information on their origin, generation methods, and scientific tools. 

\textbf{(2) Establish boundaries and strengthen training programs:} Implement clear boundaries between human-led and AI-led research processes to ensure that PhD students receive comprehensive training. Key components of doctoral education(e.g., idea testing), should prioritize human involvement to maintain high standards of scientific literacy. Additionally, guidelines should be established to prevent over-reliance on AI Scientists in PhD training, ensuring that AI tools serve as supplements rather than substitutes in the educational process.

\textbf{(3) Formulate an ethics and responsibility convention:} A global convention 
should be established to define the ethical boundaries and risk management principles for AI-driven research~\citep{huang2022overview}. All researchers and institutions utilizing AI Scientists must fully disclose the generation process, algorithmic sources, training data, and potential societal risks of their findings. Additionally, a hybrid mechanism combining automated and human-in-the-loop review should be implemented for continuous ethical oversight and risk evaluation, ensuring that AI Scientist research activities remain within socially safe boundaries~\citep{jobin2019global,khan2022ethics}. Furthermore, appropriate legislation should be developed to regulate AI Scientists by imposing strict limitations on their use for specific research purposes.

\section{Future Directions}
\label{sec:future}

This section outlines feasible pathways to bridge the current implementation capability gap of AI Scientists. Addressing foundational \textbf{Basic Abilities} is paramount. While scaling laws for pre-training and post-training \citep{kaplan2020scaling, zhang2025survey} promise progressive LLM improvements, immediate strategies like well-defined Workflows \citep{li2024autoflow, gu2024large} also can mitigate current implementation weaknesses. Structuring research processes with human-defined tools allows for guided AI execution and targeted interventions. For instance, Retrieval Augmented Generation (RAG) can counteract limitations in handling long texts or accessing current information \citep{fan2024survey, arslan2024survey}, thus expanding the knowledge scope of AI systems.

A significant challenge for sophisticated \textbf{Strategic Planning} is the immense resource consumption of RL \citep{cao2024survey}. A promising direction to alleviate this involves leveraging LLMs to simulate aspects of the environment or task execution, thereby accelerating the RL feedback loop \citep{sun2025zerosearch}. By allowing the RL agent to receive quicker, albeit potentially approximate, feedback on its actions, particularly for operations that are inherently time-consuming in the real world, the sampling efficiency may be significantly improved. This could reduce the extensive wall-clock time typically required for training robust long-horizon planning and adaptive meta-thinking capabilities in complex scientific domains.

Ensuring \textbf{Reliable Verification and Fostering Collaboration} is crucial. Standardized protocols like MCP and A2A \citep{yang2025survey, ray2025survey, hou2025model} can establish basic interoperability. A promising direction is to build modular multi-agent systems, where specialized AI agents for sub-tasks (e.g., literature review, code generation) are coordinated by a central ``Planner Agent'' trained via advanced RL, leveraging existing tools (e.g., PASA \citep{he2025pasa}) rather than reinventing capabilities. Furthermore, enhanced oversight of AI Scientist inference processes is imperative, not just to prevent benchmark ``hacking'', but also to instill ethical boundaries against unscrupulous data acquisition or other problematic behaviors.

Finally, the \textbf{Evaluation} of AI Scientists~\citep{10.1145/3641289} must evolve towards a holistic, coarse-grained paradigm reflecting real-world scientific discovery's multifaceted nature. Scientific breakthroughs involve both practical utility and novelty. Thus, evaluation frameworks should go beyond single-metric optimization, adopting multi-objective criteria that assess performance gain, originality, experimental rigor, and communication clarity. This multi-faceted approach will offer a more accurate measure of an AI Scientist's true contribution, guiding development toward impactful scientific exploration.

\section{Conclusion}

The rise of AI Scientists marks a paradigm shift in scientific discovery, with large language models (LLMs) now driving the workflow from idea generation to experiment execution. Recent systems have shown promise, producing research accepted at ICLR 2025 workshops and sparking discussions on the imminence of human-level AI Scientists. However, despite this progress, AI Scientists have yet to achieve breakthroughs in computer science comparable to traditional automated tools. Based on benchmark analyses and a systematic evaluation of 28 papers from five leading AI Scientist systems, we identify a core bottleneck: the inability to reliably execute and verify experiments. This implementation gap limits both scientific rigor and the quality of the research output. We analyze its root causes and call on the community to address this critical limitation.

\textbf{Alternative Views.} An alternative viewpoint suggests that AI Scientists need not pursue completely autonomous implementation capabilities in the short term, but rather facilitate human-machine collaboration as Co-scientists to assist humans. This approach avoids the deficiencies of LLMs in Dynamic Planning capabilities and Reliable Verification capabilities, instead allowing AI to focus on its strengths, such as idea generation, while humans execute the specific experimental results \citep{weng2025cycleresearcher}. If an AI system, though unable to independently complete all implementation details, can increase human scientists' efficiency tenfold, or help human scientists conceive and verify complex ideas previously beyond reach, then it undoubtedly also qualifies as a successful ``collaborative scientist.''

\section*{Acknowledgements}
\label{sec:acknowledgements}

The genesis of this position paper traces back to the insightful discussions and interactions at the AI Co-scientist Discussion held in conjunction with ICLR 2025 on April 26, 2024 \footnote{\url{https://ai-researcher.net/social-iclr-2025}}. We extend our sincere gratitude to the invited speakers, including Chenglei Si, Jindong Wang, Yutaro Yamada, and David Ha, whose perspectives are invaluable. We also deeply appreciate the contributions of the more than 200 participants who engaged in the vibrant discussions on that day; many of the ideas explored in this work were sparked and refined through those collective interactions. We thank every participant for their engagement and for fostering a stimulating environment that significantly shaped our thinking.

\bibliography{main}

\begin{thebibliography}{98}
\providecommand{\natexlab}[1]{#1}
\providecommand{\url}[1]{\texttt{#1}}
\expandafter\ifx\csname urlstyle\endcsname\relax
  \providecommand{\doi}[1]{doi: #1}\else
  \providecommand{\doi}{doi: \begingroup \urlstyle{rm}\Url}\fi

\bibitem[Abramo et~al.(2019)Abramo, D’Angelo, and Reale]{abramo2019peer}
Giovanni Abramo, Ciriaco~Andrea D’Angelo, and Emanuela Reale.
\newblock Peer review versus bibliometrics: Which method better predicts the scholarly impact of publications?
\newblock \emph{Scientometrics}, 121:\penalty0 537--554, 2019.

\bibitem[AI(2025)]{UniteAI2025}
Unite AI.
\newblock Google's new ai “co-scientist” aims to accelerate scientific discovery.
\newblock \emph{Unite.AI}, Feb 2025.
\newblock URL \url{https://www.unite.ai/googles-new-ai-co-scientist-aims-to-accelerate-scientific-discovery/}.

\bibitem[Aleithan et~al.(2024)Aleithan, Xue, Mohajer, Nnorom, Uddin, and Wang]{aleithan2024swe}
Reem Aleithan, Haoran Xue, Mohammad~Mahdi Mohajer, Elijah Nnorom, Gias Uddin, and Song Wang.
\newblock Swe-bench+: Enhanced coding benchmark for llms.
\newblock \emph{arXiv preprint arXiv:2410.06992}, 2024.

\bibitem[Arslan et~al.(2024)Arslan, Ghanem, Munawar, and Cruz]{arslan2024survey}
Muhammad Arslan, Hussam Ghanem, Saba Munawar, and Christophe Cruz.
\newblock A survey on rag with llms.
\newblock \emph{Procedia Computer Science}, 246:\penalty0 3781--3790, 2024.

\bibitem[Ballon et~al.(2025)Ballon, Algaba, and Ginis]{ballon2025relationship}
Marthe Ballon, Andres Algaba, and Vincent Ginis.
\newblock The relationship between reasoning and performance in large language models--o3 (mini) thinks harder, not longer.
\newblock \emph{arXiv preprint arXiv:2502.15631}, 2025.

\bibitem[Bengio et~al.(2025)Bengio, Cohen, Fornasiere, Ghosn, Greiner, MacDermott, Mindermann, Oberman, Richardson, Richardson, et~al.]{bengio2025superintelligent}
Yoshua Bengio, Michael Cohen, Damiano Fornasiere, Joumana Ghosn, Pietro Greiner, Matt MacDermott, S{\"o}ren Mindermann, Adam Oberman, Jesse Richardson, Oliver Richardson, et~al.
\newblock Superintelligent agents pose catastrophic risks: Can scientist ai offer a safer path?
\newblock \emph{arXiv preprint arXiv:2502.15657}, 2025.

\bibitem[Bo et~al.(2024)Bo, Zhang, Dai, Feng, Wang, Li, Chen, and Wen]{bo2024reflective}
Xiaohe Bo, Zeyu Zhang, Quanyu Dai, Xueyang Feng, Lei Wang, Rui Li, Xu~Chen, and Ji-Rong Wen.
\newblock Reflective multi-agent collaboration based on large language models.
\newblock \emph{Advances in Neural Information Processing Systems}, 37:\penalty0 138595--138631, 2024.

\bibitem[Boiko et~al.(2023)Boiko, MacKnight, Kline, and Gomes]{boiko2023autonomous}
Daniil~A Boiko, Robert MacKnight, Ben Kline, and Gabe Gomes.
\newblock Autonomous chemical research with large language models.
\newblock \emph{Nature}, 624\penalty0 (7992):\penalty0 570--578, 2023.

\bibitem[Brixi et~al.(2025)Brixi, Durrant, Ku, Poli, Brockman, Chang, Gonzalez, King, Li, Merchant, et~al.]{brixi2025genome}
Garyk Brixi, Matthew~G Durrant, Jerome Ku, Michael Poli, Greg Brockman, Daniel Chang, Gabriel~A Gonzalez, Samuel~H King, David~B Li, Aditi~T Merchant, et~al.
\newblock Genome modeling and design across all domains of life with evo 2.
\newblock \emph{BioRxiv}, pages 2025--02, 2025.

\bibitem[Cao et~al.(2024)Cao, Zhao, Cheng, Shu, Chen, Liu, Liang, Zhao, Yan, and Li]{cao2024survey}
Yuji Cao, Huan Zhao, Yuheng Cheng, Ting Shu, Yue Chen, Guolong Liu, Gaoqi Liang, Junhua Zhao, Jinyue Yan, and Yun Li.
\newblock Survey on large language model-enhanced reinforcement learning: Concept, taxonomy, and methods.
\newblock \emph{IEEE Transactions on Neural Networks and Learning Systems}, 2024.

\bibitem[Cemri et~al.(2025)Cemri, Pan, Yang, Agrawal, Chopra, Tiwari, Keutzer, Parameswaran, Klein, Ramchandran, et~al.]{cemri2025multi}
Mert Cemri, Melissa~Z Pan, Shuyi Yang, Lakshya~A Agrawal, Bhavya Chopra, Rishabh Tiwari, Kurt Keutzer, Aditya Parameswaran, Dan Klein, Kannan Ramchandran, et~al.
\newblock Why do multi-agent llm systems fail?
\newblock \emph{arXiv preprint arXiv:2503.13657}, 2025.

\bibitem[Chai et~al.(2024)Chai, Herron, Cervantes, and Ghosal]{chai2024exploring}
Miaosen Chai, Emily Herron, Erick Cervantes, and Tirthankar Ghosal.
\newblock Exploring scientific hypothesis generation with mamba.
\newblock In \emph{Proceedings of the 1st Workshop on NLP for Science (NLP4Science)}, pages 197--207, 2024.

\bibitem[Chan et~al.(2024)Chan, Chowdhury, Jaffe, Aung, Sherburn, Mays, Starace, Liu, Maksin, Patwardhan, et~al.]{chan2024mle}
Jun~Shern Chan, Neil Chowdhury, Oliver Jaffe, James Aung, Dane Sherburn, Evan Mays, Giulio Starace, Kevin Liu, Leon Maksin, Tejal Patwardhan, et~al.
\newblock Mle-bench: Evaluating machine learning agents on machine learning engineering.
\newblock \emph{arXiv preprint arXiv:2410.07095}, 2024.

\bibitem[Chang et~al.(2024)Chang, Wang, Wang, Wu, Yang, Zhu, Chen, Yi, Wang, Wang, Ye, Zhang, Chang, Yu, Yang, and Xie]{10.1145/3641289}
Yupeng Chang, Xu~Wang, Jindong Wang, Yuan Wu, Linyi Yang, Kaijie Zhu, Hao Chen, Xiaoyuan Yi, Cunxiang Wang, Yidong Wang, Wei Ye, Yue Zhang, Yi~Chang, Philip~S. Yu, Qiang Yang, and Xing Xie.
\newblock A survey on evaluation of large language models.
\newblock \emph{ACM Trans. Intell. Syst. Technol.}, 15\penalty0 (3), March 2024.
\newblock ISSN 2157-6904.
\newblock \doi{10.1145/3641289}.
\newblock URL \url{https://doi.org/10.1145/3641289}.

\bibitem[Chen et~al.(2021)Chen, Tworek, Jun, Yuan, Pinto, Kaplan, Edwards, Burda, Joseph, Brockman, et~al.]{chen2021evaluating}
Mark Chen, Jerry Tworek, Heewoo Jun, Qiming Yuan, Henrique Ponde De~Oliveira Pinto, Jared Kaplan, Harri Edwards, Yuri Burda, Nicholas Joseph, Greg Brockman, et~al.
\newblock Evaluating large language models trained on code.
\newblock \emph{arXiv preprint arXiv:2107.03374}, 2021.

\bibitem[Chen et~al.(2025)Chen, Chen, Ning, Zhang, Wang, Yu, Li, Liao, Wei, Lu, Dey, Xue, Baker, Burns, Adu-Ampratwum, Huang, Ning, Gao, Su, and Sun]{chen2025scienceagentbench}
Ziru Chen, Shijie Chen, Yuting Ning, Qianheng Zhang, Boshi Wang, Botao Yu, Yifei Li, Zeyi Liao, Chen Wei, Zitong Lu, Vishal Dey, Mingyi Xue, Frazier~N. Baker, Benjamin Burns, Daniel Adu-Ampratwum, Xuhui Huang, Xia Ning, Song Gao, Yu~Su, and Huan Sun.
\newblock Scienceagentbench: Toward rigorous assessment of language agents for data-driven scientific discovery.
\newblock In \emph{The Thirteenth International Conference on Learning Representations}, 2025.
\newblock URL \url{https://openreview.net/forum?id=6z4YKr0GK6}.

\bibitem[Cortes and Lawrence(2021)]{cortes2021inconsistency}
Corinna Cortes and Neil~D Lawrence.
\newblock Inconsistency in conference peer review: Revisiting the 2014 neurips experiment.
\newblock \emph{arXiv preprint arXiv:2109.09774}, 2021.

\bibitem[Fan et~al.(2024)Fan, Ding, Ning, Wang, Li, Yin, Chua, and Li]{fan2024survey}
Wenqi Fan, Yujuan Ding, Liangbo Ning, Shijie Wang, Hengyun Li, Dawei Yin, Tat-Seng Chua, and Qing Li.
\newblock A survey on rag meeting llms: Towards retrieval-augmented large language models.
\newblock In \emph{Proceedings of the 30th ACM SIGKDD Conference on Knowledge Discovery and Data Mining}, pages 6491--6501, 2024.

\bibitem[Garikaparthi et~al.(2025)Garikaparthi, Patwardhan, Vig, and Cohan]{garikaparthi2025iris}
Aniketh Garikaparthi, Manasi Patwardhan, Lovekesh Vig, and Arman Cohan.
\newblock Iris: Interactive research ideation system for accelerating scientific discovery.
\newblock \emph{arXiv preprint arXiv:2504.16728}, 2025.

\bibitem[Ghafarollahi and Buehler(2024)]{ghafarollahi2024sciagents}
Alireza Ghafarollahi and Markus~J Buehler.
\newblock Sciagents: Automating scientific discovery through multi-agent intelligent graph reasoning.
\newblock \emph{arXiv preprint arXiv:2409.05556}, 2024.

\bibitem[Gottweis et~al.(2025)Gottweis, Weng, Daryin, Tu, Palepu, Sirkovic, Myaskovsky, Weissenberger, Rong, Tanno, et~al.]{gottweis2025towards}
Juraj Gottweis, Wei-Hung Weng, Alexander Daryin, Tao Tu, Anil Palepu, Petar Sirkovic, Artiom Myaskovsky, Felix Weissenberger, Keran Rong, Ryutaro Tanno, et~al.
\newblock Towards an ai co-scientist.
\newblock \emph{arXiv preprint arXiv:2502.18864}, 2025.

\bibitem[Gu et~al.(2024{\natexlab{a}})Gu, Wang, Zhang, and Li]{gu2024llms}
Tianyang Gu, Jingjin Wang, Zhihao Zhang, and HaoHong Li.
\newblock Llms can realize combinatorial creativity: generating creative ideas via llms for scientific research.
\newblock \emph{arXiv preprint arXiv:2412.14141}, 2024{\natexlab{a}}.

\bibitem[Gu et~al.(2024{\natexlab{b}})Gu, You, Cao, Yu, Fan, and Qian]{gu2024large}
Yang Gu, Hengyu You, Jian Cao, Muran Yu, Haoran Fan, and Shiyou Qian.
\newblock Large language models for constructing and optimizing machine learning workflows: A survey.
\newblock \emph{arXiv preprint arXiv:2411.10478}, 2024{\natexlab{b}}.

\bibitem[Guo et~al.(2025)Guo, Yang, Zhang, Song, Zhang, Xu, Zhu, Ma, Wang, Bi, et~al.]{guo2025deepseek}
Daya Guo, Dejian Yang, Haowei Zhang, Junxiao Song, Ruoyu Zhang, Runxin Xu, Qihao Zhu, Shirong Ma, Peiyi Wang, Xiao Bi, et~al.
\newblock Deepseek-r1: Incentivizing reasoning capability in llms via reinforcement learning.
\newblock \emph{arXiv preprint arXiv:2501.12948}, 2025.

\bibitem[Guo et~al.(2024)Guo, Chen, Wang, Chang, Pei, Chawla, Wiest, and Zhang]{guo2024large}
Taicheng Guo, Xiuying Chen, Yaqi Wang, Ruidi Chang, Shichao Pei, Nitesh~V Chawla, Olaf Wiest, and Xiangliang Zhang.
\newblock Large language model based multi-agents: A survey of progress and challenges.
\newblock \emph{arXiv preprint arXiv:2402.01680}, 2024.

\bibitem[He et~al.(2025)He, Huang, Feng, Lin, Zhang, Li, et~al.]{he2025pasa}
Yichen He, Guanhua Huang, Peiyuan Feng, Yuan Lin, Yuchen Zhang, Hang Li, et~al.
\newblock Pasa: An llm agent for comprehensive academic paper search.
\newblock \emph{arXiv preprint arXiv:2501.10120}, 2025.

\bibitem[Hou et~al.(2025)Hou, Zhao, Wang, and Wang]{hou2025model}
Xinyi Hou, Yanjie Zhao, Shenao Wang, and Haoyu Wang.
\newblock Model context protocol (mcp): Landscape, security threats, and future research directions.
\newblock \emph{arXiv preprint arXiv:2503.23278}, 2025.

\bibitem[Hu et~al.(2024)Hu, Fu, Wang, Wang, Li, Xu, Lu, Jin, Pan, and Lan]{hu2024nova}
Xiang Hu, Hongyu Fu, Jinge Wang, Yifeng Wang, Zhikun Li, Renjun Xu, Yu~Lu, Yaochu Jin, Lili Pan, and Zhenzhong Lan.
\newblock Nova: An iterative planning and search approach to enhance novelty and diversity of llm generated ideas.
\newblock \emph{arXiv preprint arXiv:2410.14255}, 2024.

\bibitem[Huang et~al.(2022)Huang, Zhang, Mao, and Yao]{huang2022overview}
Changwu Huang, Zeqi Zhang, Bifei Mao, and Xin Yao.
\newblock An overview of artificial intelligence ethics.
\newblock \emph{IEEE Transactions on Artificial Intelligence}, 4\penalty0 (4):\penalty0 799--819, 2022.

\bibitem[Intology(2025)]{zochi2025}
Intology.
\newblock Zochi technical report.
\newblock \emph{arXiv}, 2025.

\bibitem[Jain et~al.(2024)Jain, Han, Gu, Li, Yan, Zhang, Wang, Solar-Lezama, Sen, and Stoica]{jain2024livecodebench}
Naman Jain, King Han, Alex Gu, Wen-Ding Li, Fanjia Yan, Tianjun Zhang, Sida Wang, Armando Solar-Lezama, Koushik Sen, and Ion Stoica.
\newblock Livecodebench: Holistic and contamination free evaluation of large language models for code.
\newblock \emph{arXiv preprint arXiv:2403.07974}, 2024.

\bibitem[Jansen et~al.(2025)Jansen, Tafjord, Radensky, Siangliulue, Hope, Mishra, Majumder, Weld, and Clark]{jansen2025codescientist}
Peter Jansen, Oyvind Tafjord, Marissa Radensky, Pao Siangliulue, Tom Hope, Bhavana~Dalvi Mishra, Bodhisattwa~Prasad Majumder, Daniel~S Weld, and Peter Clark.
\newblock Codescientist: End-to-end semi-automated scientific discovery with code-based experimentation.
\newblock \emph{arXiv preprint arXiv:2503.22708}, 2025.

\bibitem[Jiang et~al.(2025)Jiang, Schmidt, Srikanth, Xu, Kaplan, Jacenko, and Wu]{jiang2025aide}
Zhengyao Jiang, Dominik Schmidt, Dhruv Srikanth, Dixing Xu, Ian Kaplan, Deniss Jacenko, and Yuxiang Wu.
\newblock Aide: Ai-driven exploration in the space of code.
\newblock \emph{arXiv preprint arXiv:2502.13138}, 2025.

\bibitem[Jimenez et~al.(2024)Jimenez, Yang, Wettig, Yao, Pei, Press, and Narasimhan]{SWEBenchCanLanguageModels}
Carlos~E Jimenez, John Yang, Alexander Wettig, Shunyu Yao, Kexin Pei, Ofir Press, and Karthik~R Narasimhan.
\newblock Swe-bench: Can language models resolve real-world github issues?
\newblock In \emph{ICLR}, 2024.

\bibitem[Jobin et~al.(2019)Jobin, Ienca, and Vayena]{jobin2019global}
Anna Jobin, Marcello Ienca, and Effy Vayena.
\newblock The global landscape of ai ethics guidelines.
\newblock \emph{Nature machine intelligence}, 1\penalty0 (9):\penalty0 389--399, 2019.

\bibitem[Jumper et~al.(2021)Jumper, Evans, Pritzel, Green, Figurnov, Ronneberger, Tunyasuvunakool, Bates, {\v{Z}}{\'\i}dek, Potapenko, et~al.]{jumper2021highly}
John Jumper, Richard Evans, Alexander Pritzel, Tim Green, Michael Figurnov, Olaf Ronneberger, Kathryn Tunyasuvunakool, Russ Bates, Augustin {\v{Z}}{\'\i}dek, Anna Potapenko, et~al.
\newblock Highly accurate protein structure prediction with alphafold.
\newblock \emph{nature}, 596\penalty0 (7873):\penalty0 583--589, 2021.

\bibitem[Kaplan et~al.(2020)Kaplan, McCandlish, Henighan, Brown, Chess, Child, Gray, Radford, Wu, and Amodei]{kaplan2020scaling}
Jared Kaplan, Sam McCandlish, Tom Henighan, Tom~B Brown, Benjamin Chess, Rewon Child, Scott Gray, Alec Radford, Jeffrey Wu, and Dario Amodei.
\newblock Scaling laws for neural language models.
\newblock \emph{arXiv preprint arXiv:2001.08361}, 2020.

\bibitem[Khan et~al.(2022)Khan, Badshah, Liang, Waseem, Khan, Ahmad, Fahmideh, Niazi, and Akbar]{khan2022ethics}
Arif~Ali Khan, Sher Badshah, Peng Liang, Muhammad Waseem, Bilal Khan, Aakash Ahmad, Mahdi Fahmideh, Mahmood Niazi, and Muhammad~Azeem Akbar.
\newblock Ethics of ai: A systematic literature review of principles and challenges.
\newblock In \emph{Proceedings of the 26th international conference on evaluation and assessment in software engineering}, pages 383--392, 2022.

\bibitem[King et~al.(2009)King, Rowland, Oliver, Young, Aubrey, Byrne, Liakata, Markham, Pir, Soldatova, et~al.]{king2009automation}
Ross~D King, Jem Rowland, Stephen~G Oliver, Michael Young, Wayne Aubrey, Emma Byrne, Maria Liakata, Magdalena Markham, Pinar Pir, Larisa~N Soldatova, et~al.
\newblock The automation of science.
\newblock \emph{Science}, 324\penalty0 (5923):\penalty0 85--89, 2009.

\bibitem[Kon et~al.(2025)Kon, Liu, Ding, Qiu, Yang, Huang, Srinivasa, Lee, Chowdhury, and Chen]{kon2025curie}
Patrick Tser~Jern Kon, Jiachen Liu, Qiuyi Ding, Yiming Qiu, Zhenning Yang, Yibo Huang, Jayanth Srinivasa, Myungjin Lee, Mosharaf Chowdhury, and Ang Chen.
\newblock Curie: Toward rigorous and automated scientific experimentation with ai agents.
\newblock \emph{arXiv preprint arXiv:2502.16069}, 2025.

\bibitem[Laban et~al.(2025)Laban, Hayashi, Zhou, and Neville]{laban2025llms}
Philippe Laban, Hiroaki Hayashi, Yingbo Zhou, and Jennifer Neville.
\newblock Llms get lost in multi-turn conversation.
\newblock \emph{arXiv preprint arXiv:2505.06120}, 2025.

\bibitem[Langley(1987)]{langley1987scientific}
P~Langley.
\newblock \emph{Scientific discovery: Computational explorations of the creative processes}.
\newblock MIT Press, 1987.

\bibitem[Li et~al.(2024{\natexlab{a}})Li, Xu, Guo, Zhao, Li, Yuan, Zhang, Jiang, Xin, Dang, et~al.]{li2024chain}
Long Li, Weiwen Xu, Jiayan Guo, Ruochen Zhao, Xingxuan Li, Yuqian Yuan, Boqiang Zhang, Yuming Jiang, Yifei Xin, Ronghao Dang, et~al.
\newblock Chain of ideas: Revolutionizing research via novel idea development with llm agents.
\newblock \emph{arXiv preprint arXiv:2410.13185}, 2024{\natexlab{a}}.

\bibitem[Li et~al.(2024{\natexlab{b}})Li, Jing, Han, Zhou, and Du]{li2024learning}
Ruochen Li, Liqiang Jing, Chi Han, Jiawei Zhou, and Xinya Du.
\newblock Learning to generate research idea with dynamic control.
\newblock \emph{arXiv preprint arXiv:2412.14626}, 2024{\natexlab{b}}.

\bibitem[Li et~al.(2024{\natexlab{c}})Li, Patel, Wang, and Du]{li2024mlr}
Ruochen Li, Teerth Patel, Qingyun Wang, and Xinya Du.
\newblock Mlr-copilot: Autonomous machine learning research based on large language models agents.
\newblock \emph{arXiv preprint arXiv:2408.14033}, 2024{\natexlab{c}}.

\bibitem[Li et~al.(2024{\natexlab{d}})Li, Xu, Mei, Hua, Rama, Raheja, Wang, Zhu, and Zhang]{li2024autoflow}
Zelong Li, Shuyuan Xu, Kai Mei, Wenyue Hua, Balaji Rama, Om~Raheja, Hao Wang, He~Zhu, and Yongfeng Zhang.
\newblock Autoflow: Automated workflow generation for large language model agents.
\newblock \emph{arXiv preprint arXiv:2407.12821}, 2024{\natexlab{d}}.

\bibitem[Liu et~al.(2023)Liu, Xia, Wang, and Zhang]{liu2023your}
Jiawei Liu, Chunqiu~Steven Xia, Yuyao Wang, and Lingming Zhang.
\newblock Is your code generated by chatgpt really correct? rigorous evaluation of large language models for code generation.
\newblock volume~36, pages 21558--21572, 2023.

\bibitem[Liu et~al.(2024{\natexlab{a}})Liu, Lu, Chen, Hu, Zhao, Lu, and Zhao]{liu2024drugagent}
Sizhe Liu, Yizhou Lu, Siyu Chen, Xiyang Hu, Jieyu Zhao, Yingzhou Lu, and Yue Zhao.
\newblock Drugagent: Automating ai-aided drug discovery programming through llm multi-agent collaboration.
\newblock \emph{arXiv preprint arXiv:2411.15692}, 2024{\natexlab{a}}.

\bibitem[Liu et~al.(2025)Liu, Yang, Xie, Ni, Gao, Li, Tang, Ouyang, Cambria, and Zhou]{liu2025researchbench}
Yujie Liu, Zonglin Yang, Tong Xie, Jinjie Ni, Ben Gao, Yuqiang Li, Shixiang Tang, Wanli Ouyang, Erik Cambria, and Dongzhan Zhou.
\newblock Researchbench: Benchmarking llms in scientific discovery via inspiration-based task decomposition.
\newblock \emph{arXiv preprint arXiv:2503.21248}, 2025.

\bibitem[Liu et~al.(2024{\natexlab{b}})Liu, Liu, Zhu, Lei, Yang, Zhang, Li, and Liu]{liu2024aigs}
Zijun Liu, Kaiming Liu, Yiqi Zhu, Xuanyu Lei, Zonghan Yang, Zhenhe Zhang, Peng Li, and Yang Liu.
\newblock Aigs: Generating science from ai-powered automated falsification.
\newblock \emph{arXiv preprint arXiv:2411.11910}, 2024{\natexlab{b}}.

\bibitem[Lu et~al.(2024)Lu, Lu, Lange, Foerster, Clune, and Ha]{lu2024ai}
Chris Lu, Cong Lu, Robert~Tjarko Lange, Jakob Foerster, Jeff Clune, and David Ha.
\newblock The ai scientist: Towards fully automated open-ended scientific discovery.
\newblock \emph{arXiv preprint arXiv:2408.06292v3}, 2024.
\newblock URL \url{https://www.arxiv.org/abs/2408.06292v3}.

\bibitem[Merchant et~al.(2023)Merchant, Batzner, Schoenholz, Aykol, Cheon, and Cubuk]{merchant2023scaling}
Amil Merchant, Simon Batzner, Samuel~S Schoenholz, Muratahan Aykol, Gowoon Cheon, and Ekin~Dogus Cubuk.
\newblock Scaling deep learning for materials discovery.
\newblock \emph{Nature}, 624\penalty0 (7990):\penalty0 80--85, 2023.

\bibitem[Mikolov et~al.(2013)Mikolov, Sutskever, Chen, Corrado, and Dean]{mikolov2013distributed}
Tomas Mikolov, Ilya Sutskever, Kai Chen, Greg~S Corrado, and Jeff Dean.
\newblock Distributed representations of words and phrases and their compositionality.
\newblock \emph{Advances in neural information processing systems}, 26, 2013.

\bibitem[Muennighoff et~al.(2025)Muennighoff, Yang, Shi, Li, Fei-Fei, Hajishirzi, Zettlemoyer, Liang, Cand{\`e}s, and Hashimoto]{muennighoff2025s1}
Niklas Muennighoff, Zitong Yang, Weijia Shi, Xiang~Lisa Li, Li~Fei-Fei, Hannaneh Hajishirzi, Luke Zettlemoyer, Percy Liang, Emmanuel Cand{\`e}s, and Tatsunori Hashimoto.
\newblock s1: Simple test-time scaling.
\newblock \emph{arXiv preprint arXiv:2501.19393}, 2025.

\bibitem[O'Neill et~al.(2025)O'Neill, Ghosal, R{\u{a}}ileanu, Walmsley, Bui, Schawinski, and Ciuc{\u{a}}]{o2025sparks}
Charles O'Neill, Tirthankar Ghosal, Roberta R{\u{a}}ileanu, Mike Walmsley, Thang Bui, Kevin Schawinski, and Ioana Ciuc{\u{a}}.
\newblock Sparks of science: Hypothesis generation using structured paper data.
\newblock \emph{arXiv preprint arXiv:2504.12976}, 2025.

\bibitem[Padigela et~al.(2025)Padigela, Shah, and Juyal]{mldevbench}
Harshith Padigela, Chintan Shah, and Dinkar Juyal.
\newblock Ml-dev-bench: Comparative analysis of ai agents on ml development workflows, 2025.
\newblock URL \url{https://arxiv.org/abs/2502.00964}.

\bibitem[Pink et~al.(2025)Pink, Wu, Vo, Turek, Mu, Huth, and Toneva]{pink2025position}
Mathis Pink, Qinyuan Wu, Vy~Ai Vo, Javier Turek, Jianing Mu, Alexander Huth, and Mariya Toneva.
\newblock Position: Episodic memory is the missing piece for long-term llm agents.
\newblock \emph{arXiv preprint arXiv:2502.06975}, 2025.

\bibitem[Pu et~al.(2025{\natexlab{a}})Pu, Feng, Grossman, Hope, Dalvi~Mishra, Latzke, Bragg, Chang, and Siangliulue]{pu2025ideasynth}
Kevin Pu, KJ~Kevin Feng, Tovi Grossman, Tom Hope, Bhavana Dalvi~Mishra, Matt Latzke, Jonathan Bragg, Joseph~Chee Chang, and Pao Siangliulue.
\newblock Ideasynth: Iterative research idea development through evolving and composing idea facets with literature-grounded feedback.
\newblock In \emph{Proceedings of the 2025 CHI Conference on Human Factors in Computing Systems}, pages 1--31, 2025{\natexlab{a}}.

\bibitem[Pu et~al.(2025{\natexlab{b}})Pu, Lin, and Chen]{pu2025piflowprincipleawarescientificdiscovery}
Yingming Pu, Tao Lin, and Hongyu Chen.
\newblock Piflow: Principle-aware scientific discovery with multi-agent collaboration, 2025{\natexlab{b}}.
\newblock URL \url{https://arxiv.org/abs/2505.15047}.

\bibitem[Qian et~al.(2024)Qian, Xie, Wang, Liu, Dang, Du, Chen, Yang, Liu, and Sun]{qian2024scaling}
Chen Qian, Zihao Xie, Yifei Wang, Wei Liu, Yufan Dang, Zhuoyun Du, Weize Chen, Cheng Yang, Zhiyuan Liu, and Maosong Sun.
\newblock Scaling large-language-model-based multi-agent collaboration.
\newblock \emph{arXiv preprint arXiv:2406.07155}, 2024.

\bibitem[Rabby et~al.(2025)Rabby, Muhammed, Mitra, and Auer]{rabby2025iterative}
Gollam Rabby, Diyana Muhammed, Prasenjit Mitra, and S{\"o}ren Auer.
\newblock Iterative hypothesis generation for scientific discovery with monte carlo nash equilibrium self-refining trees.
\newblock \emph{arXiv preprint arXiv:2503.19309}, 2025.

\bibitem[Radensky et~al.(2024)Radensky, Shahid, Fok, Siangliulue, Hope, and Weld]{radensky2024scideator}
Marissa Radensky, Simra Shahid, Raymond Fok, Pao Siangliulue, Tom Hope, and Daniel~S Weld.
\newblock Scideator: Human-llm scientific idea generation grounded in research-paper facet recombination.
\newblock \emph{arXiv preprint arXiv:2409.14634}, 2024.

\bibitem[Ray(2025)]{ray2025survey}
Partha~Pratim Ray.
\newblock A survey on model context protocol: Architecture, state-of-the-art, challenges and future directions.
\newblock \emph{Authorea Preprints}, 2025.

\bibitem[Saeedi et~al.(2025)Saeedi, Buckner, Aponte, and Aghazadeh]{saeedi2025astroagents}
Daniel Saeedi, Denise Buckner, Jose~C Aponte, and Amirali Aghazadeh.
\newblock Astroagents: A multi-agent ai for hypothesis generation from mass spectrometry data.
\newblock \emph{arXiv preprint arXiv:2503.23170}, 2025.

\bibitem[Sanyal et~al.(2025)Sanyal, Schapiro, Shashidhar, Moon, Varshney, and Hakkani-Tur]{sanyal2025spark}
Aishik Sanyal, Samuel Schapiro, Sumuk Shashidhar, Royce Moon, Lav~R Varshney, and Dilek Hakkani-Tur.
\newblock Spark: A system for scientifically creative idea generation.
\newblock \emph{arXiv preprint arXiv:2504.20090}, 2025.

\bibitem[Schmidgall et~al.(2025)Schmidgall, Su, Wang, Sun, Wu, Yu, Liu, Liu, and Barsoum]{schmidgall2025agent}
Samuel Schmidgall, Yusheng Su, Ze~Wang, Ximeng Sun, Jialian Wu, Xiaodong Yu, Jiang Liu, Zicheng Liu, and Emad Barsoum.
\newblock Agent laboratory: Using llm agents as research assistants.
\newblock \emph{arXiv preprint arXiv:2501.04227}, 2025.

\bibitem[Seo et~al.(2025)Seo, Baek, Lee, and Hwang]{seo2025paper2code}
Minju Seo, Jinheon Baek, Seongyun Lee, and Sung~Ju Hwang.
\newblock Paper2code: Automating code generation from scientific papers in machine learning.
\newblock \emph{arXiv preprint arXiv:2504.17192}, 2025.

\bibitem[Shen et~al.(2025)Shen, Li, Meng, Cai, Qi, Zhang, Xu, and Ma]{shen2025shortcutsbenchlargescalerealworldbenchmark}
Haiyang Shen, Yue Li, Desong Meng, Dongqi Cai, Sheng Qi, Li~Zhang, Mengwei Xu, and Yun Ma.
\newblock Shortcutsbench: A large-scale real-world benchmark for api-based agents, 2025.
\newblock URL \url{https://arxiv.org/abs/2407.00132}.

\bibitem[Si et~al.(2024)Si, Yang, and Hashimoto]{si2024can}
Chenglei Si, Diyi Yang, and Tatsunori Hashimoto.
\newblock Can llms generate novel research ideas? a large-scale human study with 100+ nlp researchers.
\newblock \emph{arXiv preprint arXiv:2409.04109}, 2024.

\bibitem[Si et~al.(2025)Si, Yang, and Hashimoto]{SiYH25}
Chenglei Si, Diyi Yang, and Tatsunori Hashimoto.
\newblock Can llms generate novel research ideas? {A} large-scale human study with 100+ {NLP} researchers.
\newblock In \emph{The Thirteenth International Conference on Learning Representations, {ICLR} 2025, Singapore, April 24-28, 2025}. OpenReview.net, 2025.
\newblock URL \url{https://openreview.net/forum?id=M23dTGWCZy}.

\bibitem[Siegel et~al.(2024)Siegel, Kapoor, Nagdir, Stroebl, and Narayanan]{siegel2024core}
Zachary~S Siegel, Sayash Kapoor, Nitya Nagdir, Benedikt Stroebl, and Arvind Narayanan.
\newblock Core-bench: Fostering the credibility of published research through a computational reproducibility agent benchmark.
\newblock \emph{arXiv preprint arXiv:2409.11363}, 2024.

\bibitem[Starace et~al.(2025)Starace, Jaffe, Sherburn, Aung, Chan, Maksin, Dias, Mays, Kinsella, Thompson, et~al.]{starace2025paperbench}
Giulio Starace, Oliver Jaffe, Dane Sherburn, James Aung, Jun~Shern Chan, Leon Maksin, Rachel Dias, Evan Mays, Benjamin Kinsella, Wyatt Thompson, et~al.
\newblock Paperbench: Evaluating ai's ability to replicate ai research.
\newblock \emph{arXiv preprint arXiv:2504.01848}, 2025.

\bibitem[Stokes et~al.(2020)Stokes, Yang, Swanson, Jin, Cubillos-Ruiz, Donghia, MacNair, French, Carfrae, Bloom-Ackermann, et~al.]{stokes2020deep}
Jonathan~M Stokes, Kevin Yang, Kyle Swanson, Wengong Jin, Andres Cubillos-Ruiz, Nina~M Donghia, Craig~R MacNair, Shawn French, Lindsey~A Carfrae, Zohar Bloom-Ackermann, et~al.
\newblock A deep learning approach to antibiotic discovery.
\newblock \emph{Cell}, 180\penalty0 (4):\penalty0 688--702, 2020.

\bibitem[Su et~al.(2024)Su, Chen, Tang, Zheng, Li, Yin, Ouyang, and Dong]{su2024two}
Haoyang Su, Renqi Chen, Shixiang Tang, Xinzhe Zheng, Jingzhe Li, Zhenfei Yin, Wanli Ouyang, and Nanqing Dong.
\newblock Two heads are better than one: A multi-agent system has the potential to improve scientific idea generation.
\newblock \emph{arXiv preprint arXiv:2410.09403}, 2024.

\bibitem[Sun et~al.(2025)Sun, Qiao, Guo, Fan, Hou, Jiang, Xie, Huang, and Zhang]{sun2025zerosearch}
Hao Sun, Zile Qiao, Jiayan Guo, Xuanbo Fan, Yingyan Hou, Yong Jiang, Pengjun Xie, Fei Huang, and Yan Zhang.
\newblock Zerosearch: Incentivize the search capability of llms without searching.
\newblock \emph{arXiv preprint arXiv:2505.04588}, 2025.

\bibitem[Szymanski et~al.(2023)Szymanski, Rendy, Fei, Kumar, He, Milsted, McDermott, Gallant, Cubuk, Merchant, et~al.]{szymanski2023autonomous}
Nathan~J Szymanski, Bernardus Rendy, Yuxing Fei, Rishi~E Kumar, Tanjin He, David Milsted, Matthew~J McDermott, Max Gallant, Ekin~Dogus Cubuk, Amil Merchant, et~al.
\newblock An autonomous laboratory for the accelerated synthesis of novel materials.
\newblock \emph{Nature}, 624\penalty0 (7990):\penalty0 86--91, 2023.

\bibitem[Wang et~al.(2024{\natexlab{a}})Wang, Downey, Ji, and Hope]{wang-etal-2024-scimon}
Qingyun Wang, Doug Downey, Heng Ji, and Tom Hope.
\newblock {S}ci{MON}: Scientific inspiration machines optimized for novelty.
\newblock In Lun-Wei Ku, Andre Martins, and Vivek Srikumar, editors, \emph{Proceedings of the 62nd Annual Meeting of the Association for Computational Linguistics (Volume 1: Long Papers)}, pages 279--299, Bangkok, Thailand, August 2024{\natexlab{a}}. Association for Computational Linguistics.
\newblock \doi{10.18653/v1/2024.acl-long.18}.
\newblock URL \url{https://aclanthology.org/2024.acl-long.18/}.

\bibitem[Wang et~al.(2024{\natexlab{b}})Wang, Gu, Zhang, Luo, Dai, Shen, Xie, Lin, He, and Ye]{wang2024scipip}
Wenxiao Wang, Lihui Gu, Liye Zhang, Yunxiang Luo, Yi~Dai, Chen Shen, Liang Xie, Binbin Lin, Xiaofei He, and Jieping Ye.
\newblock Scipip: An llm-based scientific paper idea proposer.
\newblock \emph{arXiv preprint arXiv:2410.23166}, 2024{\natexlab{b}}.

\bibitem[Wang et~al.(2024{\natexlab{c}})Wang, Guo, Yao, Zhang, Zhang, Wu, Zhang, Dai, Wen, Ye, et~al.]{wang2024autosurvey}
Yidong Wang, Qi~Guo, Wenjin Yao, Hongbo Zhang, Xin Zhang, Zhen Wu, Meishan Zhang, Xinyu Dai, Qingsong Wen, Wei Ye, et~al.
\newblock Autosurvey: Large language models can automatically write surveys.
\newblock \emph{Advances in Neural Information Processing Systems}, 37:\penalty0 115119--115145, 2024{\natexlab{c}}.

\bibitem[Wei et~al.(2025)Wei, Sun, Papay, McKinney, Han, Fulford, Chung, Passos, Fedus, and Glaese]{wei2025browsecomp}
Jason Wei, Zhiqing Sun, Spencer Papay, Scott McKinney, Jeffrey Han, Isa Fulford, Hyung~Won Chung, Alex~Tachard Passos, William Fedus, and Amelia Glaese.
\newblock Browsecomp: A simple yet challenging benchmark for browsing agents.
\newblock \emph{arXiv preprint arXiv:2504.12516}, 2025.

\bibitem[Weng et~al.(2023)Weng, Zhu, Xia, Li, He, Liu, Sun, Liu, and Zhao]{weng2023large}
Yixuan Weng, Minjun Zhu, Fei Xia, Bin Li, Shizhu He, Shengping Liu, Bin Sun, Kang Liu, and Jun Zhao.
\newblock Large language models are better reasoners with self-verification.
\newblock In \emph{Findings of the Association for Computational Linguistics: EMNLP 2023}, pages 2550--2575, 2023.

\bibitem[Weng et~al.(2025)Weng, Zhu, Bao, Zhang, Wang, Zhang, and Yang]{weng2025cycleresearcher}
Yixuan Weng, Minjun Zhu, Guangsheng Bao, Hongbo Zhang, Jindong Wang, Yue Zhang, and Linyi Yang.
\newblock Cycleresearcher: Improving automated research via automated review.
\newblock In \emph{The Thirteenth International Conference on Learning Representations}, 2025.
\newblock URL \url{https://openreview.net/forum?id=bjcsVLoHYs}.

\bibitem[Wu et~al.(2025{\natexlab{a}})Wu, Bai, Hu, Tu, Hee, Li, and Lee]{wu2025shifting}
Yuhao Wu, Yushi Bai, Zhiqing Hu, Shangqing Tu, Ming~Shan Hee, Juanzi Li, and Roy Ka-Wei Lee.
\newblock Shifting long-context llms research from input to output.
\newblock \emph{arXiv preprint arXiv:2503.04723}, 2025{\natexlab{a}}.

\bibitem[Wu et~al.(2025{\natexlab{b}})Wu, Wang, Du, Jegelka, and Wang]{wu2025more}
Yuyang Wu, Yifei Wang, Tianqi Du, Stefanie Jegelka, and Yisen Wang.
\newblock When more is less: Understanding chain-of-thought length in llms.
\newblock \emph{arXiv preprint arXiv:2502.07266}, 2025{\natexlab{b}}.

\bibitem[Xiang et~al.(2025)Xiang, Yan, Ouyang, Gui, and He]{xiang2025scireplicate}
Yanzheng Xiang, Hanqi Yan, Shuyin Ouyang, Lin Gui, and Yulan He.
\newblock Scireplicate-bench: Benchmarking llms in agent-driven algorithmic reproduction from research papers.
\newblock \emph{arXiv preprint arXiv:2504.00255}, 2025.

\bibitem[Xiong et~al.(2024)Xiong, Xie, Shariatmadari, Guo, Bekiranov, and Zhang]{xiong2024improving}
Guangzhi Xiong, Eric Xie, Amir~Hassan Shariatmadari, Sikun Guo, Stefan Bekiranov, and Aidong Zhang.
\newblock Improving scientific hypothesis generation with knowledge grounded large language models.
\newblock \emph{arXiv preprint arXiv:2411.02382}, 2024.

\bibitem[Yamada et~al.(2025)Yamada, Lange, Lu, Hu, Lu, Foerster, Clune, and Ha]{yamada2025ai}
Yutaro Yamada, Robert~Tjarko Lange, Cong Lu, Shengran Hu, Chris Lu, Jakob Foerster, Jeff Clune, and David Ha.
\newblock The ai scientist-v2: Workshop-level automated scientific discovery via agentic tree search.
\newblock \emph{arXiv preprint arXiv:2504.08066}, 2025.

\bibitem[Yang et~al.(2025{\natexlab{a}})Yang, Li, Yang, Zhang, Hui, Zheng, Yu, Gao, Huang, Lv, et~al.]{yang2025qwen3}
An~Yang, Anfeng Li, Baosong Yang, Beichen Zhang, Binyuan Hui, Bo~Zheng, Bowen Yu, Chang Gao, Chengen Huang, Chenxu Lv, et~al.
\newblock Qwen3 technical report.
\newblock \emph{arXiv preprint arXiv:2505.09388}, 2025{\natexlab{a}}.

\bibitem[Yang et~al.(2025{\natexlab{b}})Yang, Chai, Song, Qi, Wen, Li, Liao, Hu, Lin, Chang, et~al.]{yang2025survey}
Yingxuan Yang, Huacan Chai, Yuanyi Song, Siyuan Qi, Muning Wen, Ning Li, Junwei Liao, Haoyi Hu, Jianghao Lin, Gaowei Chang, et~al.
\newblock A survey of ai agent protocols.
\newblock \emph{arXiv preprint arXiv:2504.16736}, 2025{\natexlab{b}}.

\bibitem[Yang et~al.(2025{\natexlab{c}})Yang, Yin, Kong, Chi, Tao, Zhang, and Xu]{yang2025shennongalpha}
Zijie Yang, Yongjing Yin, Chaojun Kong, Tiange Chi, Wufan Tao, Yue Zhang, and Tian Xu.
\newblock Shennongalpha: an ai-driven sharing and collaboration platform for intelligent curation, acquisition, and translation of natural medicinal material knowledge.
\newblock \emph{Cell Discovery}, 11\penalty0 (1):\penalty0 32, 2025{\natexlab{c}}.

\bibitem[Yang et~al.(2024)Yang, Liu, Gao, Xie, Li, Ouyang, Poria, Cambria, and Zhou]{yang2024moose}
Zonglin Yang, Wanhao Liu, Ben Gao, Tong Xie, Yuqiang Li, Wanli Ouyang, Soujanya Poria, Erik Cambria, and Dongzhan Zhou.
\newblock Moose-chem: Large language models for rediscovering unseen chemistry scientific hypotheses.
\newblock \emph{arXiv preprint arXiv:2410.07076}, 2024.

\bibitem[Yang et~al.(2025{\natexlab{d}})Yang, Liu, Gao, Xie, Li, Ouyang, Poria, Cambria, and Zhou]{yang2025moosechem}
Zonglin Yang, Wanhao Liu, Ben Gao, Tong Xie, Yuqiang Li, Wanli Ouyang, Soujanya Poria, Erik Cambria, and Dongzhan Zhou.
\newblock {MOOSE}-chem: Large language models for rediscovering unseen chemistry scientific hypotheses.
\newblock In \emph{The Thirteenth International Conference on Learning Representations}, 2025{\natexlab{d}}.
\newblock URL \url{https://openreview.net/forum?id=X9OfMNNepI}.

\bibitem[Yu et~al.(2024)Yu, Hong, Cheng, Zhu, Xuan, Yao, Feng, and You]{yu2024researchtown}
Haofei Yu, Zhaochen Hong, Zirui Cheng, Kunlun Zhu, Keyang Xuan, Jinwei Yao, Tao Feng, and Jiaxuan You.
\newblock Researchtown: Simulator of human research community.
\newblock \emph{arXiv preprint arXiv:2412.17767}, 2024.

\bibitem[Yuan et~al.(2025)Yuan, Yan, Shi, Chen, Ouyang, Zhang, Bai, Qiao, and Zhou]{yuan2025dolphin}
Jiakang Yuan, Xiangchao Yan, Botian Shi, Tao Chen, Wanli Ouyang, Bo~Zhang, Lei Bai, Yu~Qiao, and Bowen Zhou.
\newblock Dolphin: Closed-loop open-ended auto-research through thinking, practice, and feedback.
\newblock \emph{arXiv preprint arXiv:2501.03916}, 2025.

\bibitem[Zhang et~al.(2025)Zhang, Lyu, Sun, Wang, Zhang, Hua, Wu, Guo, Wang, Muennighoff, et~al.]{zhang2025survey}
Qiyuan Zhang, Fuyuan Lyu, Zexu Sun, Lei Wang, Weixu Zhang, Wenyue Hua, Haolun Wu, Zhihan Guo, Yufei Wang, Niklas Muennighoff, et~al.
\newblock A survey on test-time scaling in large language models: What, how, where, and how well?
\newblock \emph{arXiv preprint arXiv:2503.24235}, 2025.

\bibitem[Zhang et~al.(2024)Zhang, Sun, Chen, Pfister, Zhang, and Arik]{zhang2024chain}
Yusen Zhang, Ruoxi Sun, Yanfei Chen, Tomas Pfister, Rui Zhang, and Sercan Arik.
\newblock Chain of agents: Large language models collaborating on long-context tasks.
\newblock \emph{Advances in Neural Information Processing Systems}, 37:\penalty0 132208--132237, 2024.

\bibitem[Zheng et~al.(2024)Zheng, Sun, Qiu, Ru, Jiayang, Li, Lin, Wang, Luo, Pan, et~al.]{zheng2024openresearcher}
Yuxiang Zheng, Shichao Sun, Lin Qiu, Dongyu Ru, Cheng Jiayang, Xuefeng Li, Jifan Lin, Binjie Wang, Yun Luo, Renjie Pan, et~al.
\newblock Openresearcher: Unleashing ai for accelerated scientific research.
\newblock \emph{arXiv preprint arXiv:2408.06941}, 2024.

\bibitem[Zhu et~al.(2025)Zhu, Weng, Yang, and Zhang]{zhu2025deepreview}
Minjun Zhu, Yixuan Weng, Linyi Yang, and Yue Zhang.
\newblock Deepreview: Improving llm-based paper review with human-like deep thinking process.
\newblock \emph{arXiv preprint arXiv:2503.08569}, 2025.

\end{thebibliography}

\appendix
\section{Sampling Time Calculation for Different Types of AI Agents}
\label{appendix:1}
We referenced existing literature \citep{guo2025deepseek,yang2025qwen3,muennighoff2025s1} and our experience to estimate the sampling time potentially required for different types of AI agents trained via reinforcement learning, as illustrated in Figure~\ref{fig:sampling_time_estimation}. Using a hypothetical 671B parameter LLM (similar to Deepseek-R1) running on 8 H100 cards (assuming 40 tokens generated per second), the pure inference time $T_{infer\_R}$ for a typical arithmetic reasoning task (generating approximately 10,000 tokens of reasoning content) might be around 250 seconds. For an AI Web Agent, the task might include generating approximately 8,000 tokens of instructions and reports ($T_{infer\_WA} \approx 200s$), interspersed with approximately 20 API calls for information search (assuming each search and processing takes $T_{search\_API} = 10s$, totaling $T_{search\_total} = 20 \times 10s = 200s$), and potentially requiring reading and comprehension of up to 400,000 tokens of web content (assuming reading and comprehension time $T_{read\_WA} \approx 200s$). The total sampling time is: $T_{sample\_WA} \approx T_{infer\_WA} + T_{search\_total} + T_{read\_WA} \approx 600s$.

In contrast, an AI Scientist executing end-to-end scientific discovery tasks has complexity and interaction requirements far exceeding the previous two types. We roughly estimate it might need to generate over 100,000 tokens of content (for example, operational and experimental code approximately 50,000 tokens ($T_{infer\_code} \approx 1250s$), research paper writing approximately 30,000 tokens ($T_{infer\_paper} \approx 750s$), reviewing and understanding relevant literature approximately 20,000 tokens ($T_{infer\_lit} \approx 500s$)), with pure LLM inference time for just this portion being $T_{infer\_CS} = T_{infer\_code} + T_{infer\_paper} + T_{infer\_lit} \approx 2500s$. More critically, the "implementation" process of an AI Scientist, such as code writing, debugging, compiling, running experiments, and data analysis, is highly asynchronous and time-consuming. Assuming a rapid research code operation and experimental cycle (from writing to obtaining preliminary results) requires an average of $T_{op\_code} \approx 12 \text{ hours} = 43200s$, while in-depth literature research and analysis might require $T_{op\_lit} \approx 20 \text{ minutes} = 1200s$. Therefore, the total estimated sampling time to complete a relatively complete scientific exploration loop would be $T_{sample\_CS} = T_{infer\_CS} + T_{op\_code} + T_{op\_lit} \approx 2500s + 43200s + 1200s \approx 46900s$. As intuitively demonstrated in Figure~\ref{fig:sampling_time_estimation}, the sampling time required for an AI Scientist (approximately 46,000 seconds) far exceeds that of an AI Reasoner (approximately 250 seconds) and an AI Web Agent (approximately 700 seconds). Notably, AI Reasoners can typically rapidly generate large quantities of training samples through batch generation in parallel, whereas each implementation step of an AI Scientist (especially parts involving code execution and experiment waiting) is almost entirely asynchronous, and requires exclusive computational resources or experimental equipment for learning and feedback collection during operations. Consequently, in actual reinforcement learning training processes, the disparity in real training duration between AI Scientists and the former two types will be even more pronounced.

For the calculation of human duration, we referenced existing metrics. For instance, for reasoning tasks, we referred to the human time from the International Mathematical Olympiad, which is approximately 1.5 hours per problem. For Web Agent tasks, we adopted the average human problem-solving time from BrowseComp \citep{wei2025browsecomp} (2 hours) as the human standard. For Scientist tasks, although each paper often requires months of collaborative work by multiple people, for ease of calculation, we used the human duration of 48 hours from PaperBench \citep{starace2025paperbench} for statistics; however, even under these conditions, humans achieve a success rate of less than 50\%.

\section{Regarding the statistics for the papers}
\label{sec:statistics}
We have conducted a comprehensive search on arXiv to gather relevant publications in the AI Scientist field. This collection includes a series of papers from August 2024 to April 2025 for methods or systems, which are cited in Table \ref{paper_count}. It indicates that, to date, a significant number of papers have focused on Idea Generation tasks, often without concrete implementations.

Nevertheless, an encouraging trend has emerged since early 2025. As illustrated in Figure~\ref{fig:difference_im}, implementation-focused research has demonstrated stronger growth momentum, with incremental growth nearly matching that of non-implementation studies by Spring 2025. This suggests the community is beginning to recognize the critical importance of implementation capabilities for developing truly effective AI Scientists—moving beyond theoretical constructs toward practical systems capable of reliable execution.
\begin{table}[t]

\centering
\caption{Timeline of AI Scientist Ideas and Code Implementations by Month}
\label{paper_count}
\footnotesize
\resizebox{\textwidth}{!}{%
\begin{tabular}{cp{2cm}p{2cm}p{2cm}p{2cm}p{2cm}p{2cm}p{2cm}p{2cm}p{2cm}}

\toprule

& \textbf{2024-08} & \textbf{2024-09} & \textbf{2024-10} & \textbf{2024-11} & \textbf{2024-12} & \textbf{2025-01} & \textbf{2025-02} & \textbf{2025-03} & \textbf{2025-04} \\

\hline

\textbf{w/o Exp} & \citep{zheng2024openresearcher} & \citep{ghafarollahi2024sciagents}, \citep{radensky2024scideator} & \citep{pu2025ideasynth}, \citep{yang2024moose}, \citep{su2024two}, \citep{li2024chain}, \citep{hu2024nova}, \citep{liu2025researchbench}, \citep{wang2024scipip} & \citep{weng2025cycleresearcher}, \citep{xiong2024improving} & \citep{gu2024llms}, \citep{li2024learning}, \citep{yu2024researchtown} & & \citep{gottweis2025towards} & \citep{rabby2025iterative}, \citep{saeedi2025astroagents} & \citep{o2025sparks}, \citep{garikaparthi2025iris}, \citep{sanyal2025spark} \\

\rowcolor[rgb]{ .949,  .949,  .949}
\textbf{w/ Exp} & \citep{lu2024ai}, \citep{li2024mlr} & & & \citep{liu2024aigs}, \citep{liu2024drugagent} & & \citep{yuan2025dolphin}, \citep{schmidgall2025agent} & \citep{jiang2025aide}, \citep{kon2025curie} & \citep{schmidgall2025agent}, \citep{jansen2025codescientist} & \citep{yamada2025ai}, \citep{seo2025paper2code} \\

\toprule

\end{tabular}}

\end{table}

\appendix
\end{document}